\documentclass[12pt]{article}

\usepackage{graphicx}
\usepackage{amssymb}
\usepackage{amsmath}
\usepackage{amsbsy}
\usepackage{setspace}
\usepackage[margin=1in]{geometry}
\usepackage{natbib}
\usepackage{bm}
\usepackage{textcomp}
\usepackage{xcolor}

\usepackage{algorithm}
\usepackage{algorithmic}
\usepackage{makecell}
\usepackage{listings}

\newcommand{\INPUT}{\item[\textbf{Input:}]} 
\newcommand{\OUTPUT}{\item[\textbf{Output:}]}

\usepackage{amsthm}
\usepackage{enumitem}
\usepackage{longtable,threeparttablex,booktabs}
\usepackage{mathtools}
\usepackage{lscape}
\usepackage{subcaption}
\usepackage{multirow}
\usepackage{lineno}
\usepackage{comment}
\usepackage{url}
\usepackage[colorlinks=true,linkcolor=black,citecolor=black,urlcolor=black]{hyperref}

\numberwithin{equation}{section}
\newtheorem{theorem}{Theorem}[section]

\newtheorem{proposition}{Proposition}[section]
\newtheorem{lemma}{Lemma}

\newtheorem{assumption}{Assumption}[section]

\def\underwiggle 1{
\ifmmode\setbox\TempBox=\hbox{$ 1$}\else\setbox\TempBox=\hbox{
1}\fi \setbox\TempBoxA=\hbox to \wd\TempBox{\hss\char'176\hss}
\rlap{\copy\TempBox}\smash{\lower9pt\hbox{\copy\TempBoxA}} }

\graphicspath{{./Figures/}}

\title{Bias-Corrected Data Synthesis for Imbalanced Learning}
\author{Pengfei Lyu$^1$, ~ Zhengchi Ma$^2$, ~ Linjun Zhang$^3$, ~ and ~ Anru R. Zhang$^{1,4}$\thanks{Corresponding author: \href{mailto: anru.zhang@duke.edu}{anru.zhang@duke.edu}}}
\date{}

\begin{document}
\maketitle
\footnotetext[1]{Department of Biostatistics \& Bioinformatics, Duke University}
\footnotetext[2]{Department of Electrical \& Computer Engineering, Duke University}
\footnotetext[3]{Department of Statistics, Rutgers University}
\footnotetext[4]{Department of  Computer Science, Duke University}
\begin{abstract}
Class imbalance complicates probabilistic classification because standard training objectives emphasize majority-class performance. Synthetic oversampling can reduce imbalance, but discrepancies between the synthetic and target minority distributions may bias the fitted classifier, especially because synthetic samples depend on the observed data. We propose a bias-correction procedure that estimates the generator-induced loss discrepancy from a held-out subset of majority observations and transfers this correction to the minority class under a uniform bias-transfer condition. We establish finite-sample bounds for bias transfer and for the excess balanced risk of the resulting empirical risk minimizer, and characterize a regime in which SMOTE induces non-negligible loss bias. The framework can also be implemented in imbalanced multi-task learning and propensity-score estimation, with details provided in the Supplementary Material. Real data analyses show that the correction is most useful when synthetic distortion is appreciable.
\end{abstract}
\section{Introduction}\label{sec:intro}
Imbalanced classification is a fundamental challenge in modern machine learning, arising when the number of observations in one class significantly exceeds that in another class. This issue is prevalent in diverse applications, including detecting rare diseases in medical diagnosis \citep{rajkomar2019machine, faviez2020diagnosis}, fraud detection \citep{subudhi2018effect}, anomaly detection in industrial systems \citep{kong2020improving}, and cybersecurity \citep{sarker2019machine}. Traditional classification algorithms often perform poorly under such an imbalance, as they tend to be biased towards the majority class, leading to suboptimal sensitivity and an increased risk of overlooking critical minority instances. 

A common strategy for addressing such challenges in imbalanced classification is data augmentation, which aims to rebalance the samples in different classes by artificially modifying or expanding the training dataset. Resampling-based approaches include undersampling (removing samples from the majority class)  and oversampling (expanding the minority class). Undersampling techniques, such as Tomek's links \citep{tomek2010generalization} and cluster centroid \citep{lemaavztre2017imbalanced}, often suffer from information loss due to discarding potentially informative majority samples. In contrast, oversampling is typically preferred, and various methods have been proposed to enrich the minority class. The reweighting procedure, which assigns higher weights to the minority samples, is equivalent to oversampling by replicating the minority samples. While bootstrap \citep{efron1994introduction} is a widely used resampling method in statistics, its na\"{i}ve application in oversampling may be sensitive to outliers and may introduce variance inflation. The Gaussian mixture model \citep{mclachlan2000finite} is a classical tool for modeling heterogeneous populations through parametric mixture distributions. Perturbation-based sampling schemes are also used to address class imbalance; see \citet{he2009learning} for a review. Both approaches provide convenient modeling abstractions, though they rely on simplified assumptions and may not fully capture the complexity of real-world data.

Among oversampling methods, the Synthetic Minority Oversampling TEchnique (SMOTE, \citet{chawla2002smote}) has been especially influential. SMOTE generates synthetic samples by interpolating between minority samples and has inspired numerous variants, such as Borderline-SMOTE \citep{han2005borderline}, ADASYN \citep{he2008adasyn}, and safe-level-SMOTE \citep{bunkhumpornpat2009safe}, which aim to better capture the geometry of the data and concentrate synthetic sample generation near the decision boundary where classification is challenging. For a comprehensive review of resampling techniques in imbalanced settings, see \citet{mohammed2020machine}.

Beyond empirical success, recent theoretical studies have examined the statistical properties of synthetic procedures and their impact on classification risk. For example, \citet{elreedy2024theoretical} and \citet{sakhowe2024do} separately derive the probability distribution of SMOTE-generated synthetic samples, with the latter further proving that the synthetic density function vanishes near the boundary of the minority support. Another widely used augmentation method is Mixup \citep{zhang2017mixup}, which generates new samples by convex combinations of covariates and their labels. In \citet{xia2026classification}, imbalanced classification is viewed through a transfer-learning lens under label shift, and regimes are identified in which oversampling can be detrimental. From a learning-theoretic perspective, uniform concentration bounds are established in \citet{ahmad2025concentration} that link the empirical risk computed on synthetic samples to the true minority risk, yielding excess-risk guarantees for kernel-based classifiers. Theoretical results for Mixup include robustness against adversarial attacks and improved generalization by reducing overfitting \citep{zhang2020does}, as well as conditions under which Mixup helps reduce calibration errors \citep{thulasidasan2019mixup, zhang2022and}. 

Some recent work suggests that resampling/upsampling does not always improve performance under class imbalance and can even be counterproductive in certain prediction settings. Large empirical comparisons find that the benefits of common sampling strategies are highly dataset- and model-dependent, with no uniformly best choice across imbalance regimes \citep{newaz2022empirical}. In clinical risk prediction with logistic regression, it is shown that oversampling and SMOTE can substantially degrade probability calibration without improving discrimination, and that comparable operating points can often be achieved via threshold adjustment \citep{van2022harm}.  \citet{assunccao2024augmentation} argue, under a calibration-centric perspective, that adjusting the decision cutoff may recover much of the benefit typically attributed to augmentation. Motivated by the dependence introduced when synthetic samples are generated from the training data, \citet{tian2026conditional} propose a sample-splitting framework in which one subset is used to generate synthetic samples and an independent subset is used for training. Nevertheless, a fundamental question remains: under what conditions do synthetic procedures improve classification, and how can potential adverse effects be avoided? In this work, we aim to address this question by developing a principled framework to analyze and correct the distributional discrepancy introduced by synthetic data.

We summarize our contributions as follows:

{\bf Bias-corrected synthetic data augmentation for imbalanced classification.} We develop a bias-correction methodology that estimates a generator-induced loss discrepancy from majority-class data and transfers the resulting correction to the minority class. This construction provides a computable surrogate for the loss that would be obtained from an independently balanced sample, up to a bias-transfer error.

{\bf Theoretical guarantees and error bounds.} We provide finite-sample bounds for transferring the majority loss discrepancy to the minority class and for the excess balanced risk of the bias-corrected empirical risk minimizer. We also characterize conditions under which SMOTE introduces a non-negligible minority loss bias. Together, these results clarify how sampling fluctuations and structural bias-transfer error determine the quality of the correction.

{\bf Framework and practical validation.} The correction can be combined with several synthetic generators and can be implemented in multi-task learning and propensity-score estimation. Real-data analyses show that its benefit grows as generator-induced distortion becomes more pronounced. Additional simulations and data analysis are provided in the supplementary materials.

\section{Methodology}\label{sec:method}
We begin by introducing the setting for binary classification with imbalanced data. Suppose that the training data consist of $n$ independent and identically distributed (i.i.d.) samples $(\bm X_i, Y_i)_{i=1}^n$, where $\bm X_i \in \mathcal X \subseteq \mathbb R^d$ is a $d$-dimensional covariate vector and $Y_i \in \{0, 1\}$ represents the class label. Assume that the $Y_i$ follow a Bernoulli distribution with $\pi_1 = \mathbb P(Y_i = 1)$ and $\pi_0 = \mathbb P(Y_i = 0)$. In the imbalanced setting, we assume $0 < \pi_1 < 1/2$, so that the class $Y=1$ is underrepresented. For convenience, we refer to $Y=1$ as the minority class and $Y=0$ as the majority class. Let $n_1 = \sum_{i=1}^n Y_i$ and $n_0 = n-n_1$ denote the respective sample sizes, with $n_1\ll n_0$ with high probability. Without loss of generality, we assume that the samples are ordered such that $Y_1 = \dots = Y_{n_1} = 1$ and $Y_{n_1+1} = \dots = Y_n = 0$. We also assume that $\bm X\mid (Y = 1) \sim \mathcal P_1$ and $\bm X\mid (Y=0) \sim \mathcal P_0$, where $\mathcal P_1, \mathcal P_0$ represent the class-conditional distributions.


\subsection{Bias Correction with Synthetic Data}

Our goal is to learn a prediction function $f:\mathcal X\to(0,1)$ that performs well under a balanced class-conditional risk. Our framework focuses on risk minimization with respect to a general loss function and does not impose a specific classification or decision rule. In particular, the learned function $f$ can be combined with any downstream classification rule depending on the application, such as thresholding or cost-sensitive decisions. One widely used example is to obtain the prediction via $\widehat Y_{n+1} = I(f(\bm X_{n+1}) \geq 1/2)$ where $I(\cdot)$ denotes the indicator function taking values in $\{0, 1\}$. Let $\mathcal F$ be a prescribed class of measurable prediction functions
$f:\mathcal X\to(0,1)$. Its boundedness and complexity conditions are specified in Assumptions~\ref{asmp:bounded} and~\ref{asmp:complexity}. Denote the corresponding loss function as a binary cross-entropy loss, for example,  $\ell_f(\bm X, Y) = -Y\log f(\bm X) - (1-Y)\log (1-f(\bm X))$. Throughout this paper, the proposed correction is formulated at the loss and risk level and is agnostic to the specific classification architecture. The bias-corrected objective can be implemented with classifiers trained via empirical risk minimization, including logistic regression and neural networks with cross-entropy loss, as well as margin-based methods such as calibrated support vector machines and boosting. The theoretical guarantees in Section~\ref{sec_theory}, however, apply to prediction-function classes satisfying the structural bias-transfer conditions introduced there.

\begin{figure}[htbp]
    \centering
    \includegraphics[width=0.8\linewidth]{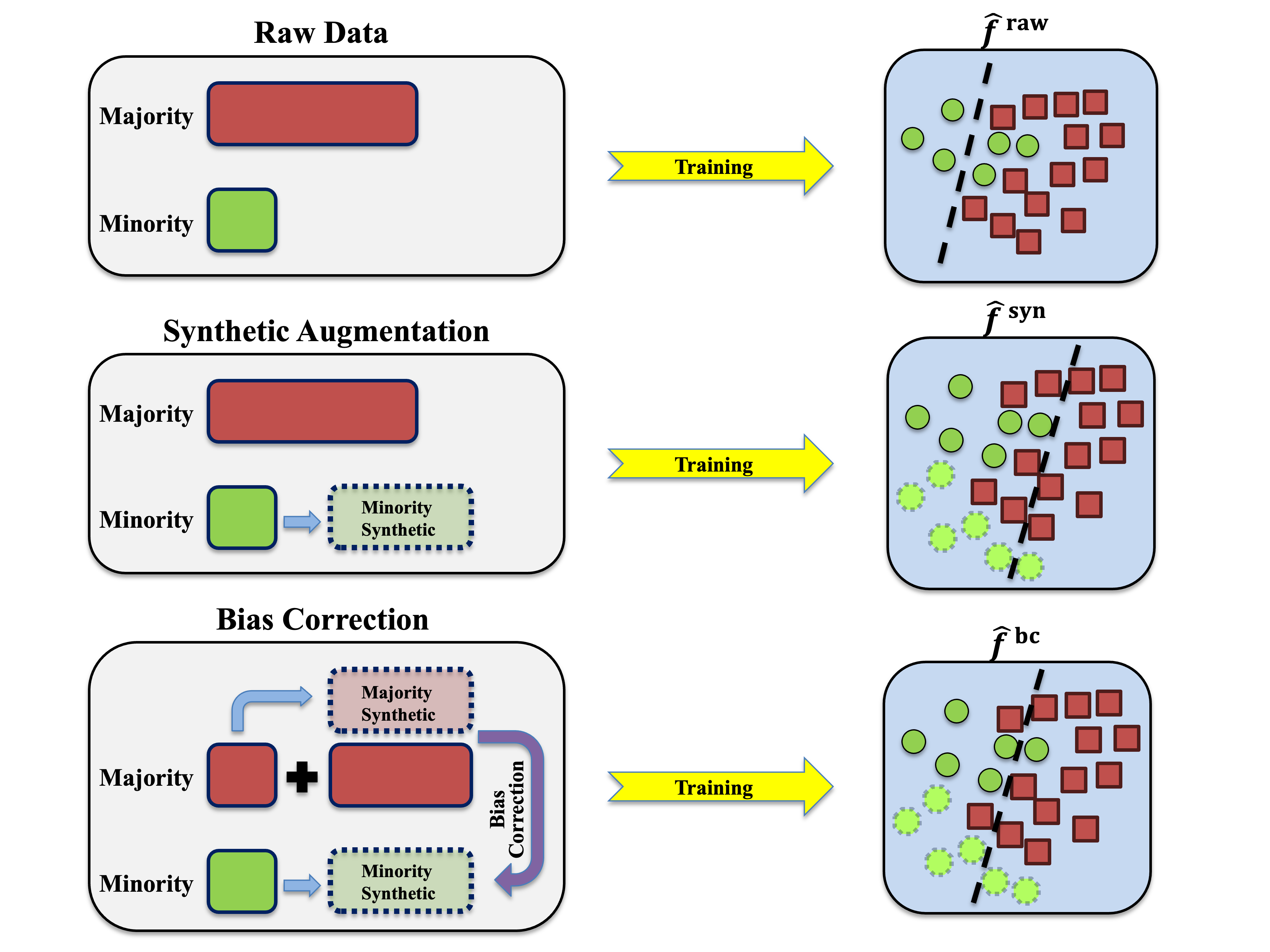}
    \caption{Flow diagram showing imbalanced observed data, synthetic minority augmentation, majority-based bias measurement, and the final bias-corrected learning procedure.}
    \label{fig_BC_illu}
\end{figure}

With the raw data $(\bm X_i, Y_i)_{i=1}^n$, we can simply train the prediction function by minimizing the empirical loss function $L^{\rm raw}$:
\begin{align}
    \label{eq_raw_loss}
    L^{\rm raw}(f) = \frac{1}{n}\sum_{i=1}^n \ell_f(\bm X_i, Y_i).
\end{align}
In the case that $n_1 \ll n_0$, a trivial guess that all samples are from the majority group will result in accuracy as high as $n_0/n$, which is close to $1$, but it does not provide information from the data. To deal with this problem, an intuitive way is to make the data balanced by adding synthetic data samples to the minority group. Assume that we have $\tilde{n}_1$ synthetic samples for the minority group: $(\tilde{\bm X}^{(1)}_i, \tilde Y_i^{(1)})_{i=1}^{\tilde{n}_1}$, where $\tilde Y_i^{(1)} = 1$ for $i = 1,\ldots, \tilde{n}_1$ with $\tilde n_1 \approx n_0 - n_1$. By equally treating the synthetic and raw samples, we can run the algorithm by minimizing the synthetic-augmented loss function $L^{\rm syn}$:
\begin{align}
    \label{eq_syn_loss}
    L^{\rm syn}(f) = \frac{1}{n+\tilde{n}_1}\bigg(\sum_{i=1}^n \ell_f(\bm X_i, Y_i) + \sum_{i=1}^{\tilde{n}_1}\ell_f(\tilde{\bm X}_i^{(1)}, \tilde Y_i^{(1)})\bigg).
\end{align}
See Figure~\ref{fig_BC_illu} for the illustration of the imbalanced learning based on raw data and synthetic augmentation.
By introducing $\tilde n_1$ synthetic samples from the minority group, $L^{\rm syn}$ is a loss function from a ``balanced'' dataset, especially compared with $L^{\rm raw}$.
While $L^{\rm syn}$ can improve prediction accuracy, a concern arises when the synthetic data fail to recover the minority distribution $\mathcal P_1$. This induces a discrepancy between the expected losses under $\mathcal P_1$ and $\tilde{\mathcal P}_1$, defined as follows:
\begin{align}
    \label{eq_population_bias_1}
    \Delta_1 = \mathbb E_{\bm X\sim\mathcal P_1}\{\ell_f(\bm X, 1)\} - \mathbb E_{\tilde{\bm X}\sim\tilde{\mathcal P}_1}\{\ell_f(\tilde{\bm X}, 1)\}.
\end{align}

Since the sample size in the minority class is limited, and the synthetic samples $(\tilde{\bm X}_i^{(1)})_{i=1}^{\tilde n_1}$ are generated conditionally on the observed samples $(\bm X_i)_{i=1}^{n_1}$, a direct comparison between the empirical losses $n_1^{-1}\sum_{i=1}^{n_1}\ell_f(\bm X_i, 1)$ and $\tilde n_1^{-1}\sum_{i=1}^{\tilde n_1}\ell_f(\tilde{\bm X}_i, 1)$ leads to variance inflation and induces a complicated dependence structure in the resulting difference.
To overcome the two challenges, consider $n_1^*$ satisfying $n_1^* \approx n_0 - n_1$. Suppose, hypothetically, that we had $n_1^*$ i.i.d. unobserved samples from the minority group $(\bm X_i^*, Y_i^*)_{i=1}^{n_1^*}$ independent of the observed data, where $Y_i^* = 1$ and $\bm X_i^*\mid (Y_i^*=1)\sim \mathcal P_1$. By introducing the unobserved minority samples, the total dataset $(\bm X_i, Y_i)_{i=1}^n$ and $(\bm X_i^*, Y_i^*)_{i=1}^{n_1^*}$ is roughly balanced. Denote the sample bias caused by the synthetic data as
\begin{align}
    \label{eq_sample_bias_1}
    \widehat\Delta_1 = \frac{1}{n_1^*}\sum_{i=1}^{n_1^*}\ell_f(\bm X_i^*, Y_i^*) - \frac{1}{\tilde{n}_1}\sum_{i=1}^{\tilde{n}_1}\ell_f(\tilde{\bm X}_i^{(1)}, \tilde Y_i^{(1)}).
\end{align}
Suppose we have the balanced data including observed dataset $(\bm X_i, Y_i)_{i=1}^n$ and unobserved dataset $(\bm X_i^*, Y_i^*)_{i=1}^{n_1^*}$. The empirical loss function for the balanced dataset is
\begin{align}
    L^{\rm bal}(f) =& \frac{1}{n+n_1^*} \left\{\sum_{i=1}^n \ell_f(\bm X_i, Y_i) + \sum_{i=1}^{n_1^*}\ell_f(\bm X_i^*, Y_i^*)\right\}
    \notag
    \\
    =& \frac{1}{n+n_1^*} \left\{\sum_{i=1}^n \ell_f(\bm X_i, Y_i) + n_1^*\cdot\left(\frac{1}{\tilde{n}_1}\sum_{i=1}^{\tilde{n}_1}\ell_f(\tilde{\bm X}_i^{(1)}, \tilde Y_i^{(1)}) + \widehat\Delta_1\right)\right\}.
    \label{eq_bal_loss}
\end{align}
Since $(\bm X_i^*, Y_i^*)_{i=1}^{n_1^*}$ are not observable, it is impossible to calculate $\widehat\Delta_1$ with the raw data. As formalized through explicit assumptions introduced later, we can estimate the bias from the available data in the majority group as follows. 
First, randomly partition the majority indices into a generation subgroup $\mathcal S_{0g}$ and a correction subgroup $\mathcal S_{0c}$ with corresponding sizes $n_{0g}$ and $n_{0c}$, respectively. In practice, we recommend choosing $n_{0g} \approx n_1$ and therefore $n_{0c} \approx n_0-n_1$. Matching the two generator-training sample sizes helps make the generator-induced distortions in the two classes comparable, which supports bias transfer. Next, using samples from the generation subgroup $(\bm X_i, Y_i)_{i\in \mathcal S_{0g}}$, generate $\tilde{n}_0$ synthetic samples $(\tilde{\bm X}_i^{(0)}, \tilde Y_i^{(0)})_{i=1}^{\tilde{n}_0}$ by the same synthetic generator, where $\tilde n_0 \approx n_0-n_1$ and $\tilde Y_i^{(0)} = 0$ for $i = 1, \ldots, \tilde{n}_0$. Then consider the population bias of the loss function from the true majority distribution $\mathcal P_0$ to the synthetic majority distribution $\tilde{\mathcal P}_0$ by
\begin{align}
    \label{eq_population_bias_0}
    \Delta_0 = \mathbb E_{\bm X\sim\mathcal P_0}\{\ell_f(\bm X, 0)\} - \mathbb E_{\tilde{\bm X}\sim\tilde{\mathcal P}_0}\{\ell_f(\tilde{\bm X}, 0)\}.
\end{align}
Finally, obtain the sample loss bias for the majority group using the majority synthetic samples and the correction subsamples by
\begin{align}
    \label{eq_sample_bias_0}
    \widehat\Delta_0 = \frac{1}{n_{0c}}\sum_{i\in \mathcal S_{0c}}\ell_f(\bm X_i, Y_i) - \frac{1}{\tilde{n}_0}\sum_{i=1}^{\tilde{n}_0}\ell_f(\tilde{\bm X}_i^{(0)}, \tilde Y_i^{(0)}).
\end{align}
Note that all elements for calculating $\widehat\Delta_0$ are available, and $\widehat\Delta_0$ is constructed in the same way as $\widehat\Delta_1$. The unobservable minority bias can be approximated by the majority bias when the generator-induced loss discrepancies are transferable across the two classes, as formalized in Assumption~\ref{asmp:bias_transfer}. Assumption~\ref{asmp_transformation} subsequently provides a stronger transformation-based sufficient condition, but this transformation structure is not required for implementing the proposed correction. To obtain a computable surrogate of the balanced loss, we further assume that the number of unobserved minority samples $n_1^*$ is matched by the number of generated synthetic minority samples, i.e., $n_1^*\approx\tilde n_1$. Thus, by modifying the balanced loss $L^{\rm bal}(f)$ in (\ref{eq_bal_loss}), we propose the following bias-correction loss function $L^{\rm bc}(f)$:
\begin{equation}
    \begin{split}
        L^{\rm bc}(f) = \frac{1}{n+\tilde{n}_1}\bigg[\sum_{i=1}^n \ell_f(\bm X_i, Y_i) + \tilde{n}_1 \cdot \bigg\{\frac{1}{\tilde{n}_1}\sum_{i=1}^{\tilde{n}_1}\ell_f(\tilde{\bm X}_i^{(1)}, \tilde Y_i^{(1)})
        + \widehat\Delta_0 
        \bigg\}\bigg].
    \end{split}
    \label{eq_bc_loss}
\end{equation}
The majority bias correction term $\widehat\Delta_0$ captures the loss function bias induced by the discrepancy between the true and synthetic distributions from the majority group. Under the structural bias-transfer conditions introduced in Section~\ref{sec_theory}, $\widehat\Delta_0$ approximates $\widehat\Delta_1$, the loss function bias from the minority group, up to a fixed bias-transfer error and sampling fluctuations. With this property, the term $\frac{1}{\tilde{n}_1}\sum_{i=1}^{\tilde n_1}\ell_f(\tilde{\bm X}_i^{(1)}, \tilde Y_i^{(1)}) + \widehat\Delta_0$ serves as a corrected surrogate for the average loss from the unobserved minority samples. Consequently, the bias-corrected loss $L^{\rm bc}$ provides a computable surrogate for the loss from a roughly balanced dataset. Finally, we can find the prediction function by minimizing the bias-corrected loss:
\begin{align}
    \label{eq_f_bc}
    \widehat f^{\rm bc} = \underset{f\in\mathcal F}{\arg\min}\: L^{\rm bc} (f).
\end{align}
The process of bias correction for imbalanced classification is summarized in Algorithm~\ref{algo_bc}. See Figure~\ref{fig_BC_illu} for an illustration.

\begin{algorithm}[htbp]
\caption{Bias Correction for Imbalanced Classification}
\label{algo_bc}
\begin{algorithmic}[1]
    \INPUT Imbalanced data $(\bm X_i, Y_i)_{i=1}^n$, prediction function class $\mathcal F$, loss function $\ell_f$, synthetic generator $\mathcal G$, minority synthetic size $\tilde n_1$, majority synthetic size $\tilde n_0$ and generation size $n_{0g}$.    
    \STATE Minority augmentation: Generate $\tilde n_1$ synthetic minority samples $(\tilde{\bm X}_i^{(1)})_{i=1}^{\tilde n_1}$ via generator $\mathcal G$ trained on the minority group $(\bm X_i)_{Y_i=1}$.
    \STATE Partition the majority index set $\mathcal S_0 = \{i : Y_i = 0\}$ into a generation set $\mathcal S_{0g}$ and a correction set $\mathcal S_{0c}$ with corresponding sizes $n_{0g}$ and $n_{0c} = n_0 - n_{0g}$.
    \STATE Generate $\tilde n_0$ synthetic majority samples $(\tilde{\bm X}_i^{(0)})_{i=1}^{\tilde n_0}$ via generator $\mathcal G$ trained on the generation set $(\bm X_i)_{i \in \mathcal S_{0g}}$.
    \STATE Compute the empirical majority bias $\widehat\Delta_0$
    according to Equation~(\ref{eq_sample_bias_0}).
    \STATE Form the bias-corrected loss $L^{\rm bc}(f)$
    as defined in Equation~(\ref{eq_bc_loss}).
    \STATE Obtain the predictor $\widehat f^{\rm bc}$ by \eqref{eq_f_bc}.
    \OUTPUT The prediction function $\widehat f^{\rm bc}: \mathbb R^d\to(0, 1)$.
\end{algorithmic}
\end{algorithm}

\subsection{Applications}
Our bias correction framework can be extended to various scenarios with imbalanced data, including multi-task learning and causal inference. 

In imbalanced multi-task learning, the correction can be applied task by task to obtain bias-corrected coefficient estimators. Aggregating these estimators into a coefficient matrix then yields an estimate of the shared representation through its leading left singular subspace. The complete construction and implementation details are provided in Section~A of the Supplementary Material.

In causal inference with observational data, the treatment group is often underrepresented compared with the control group. Synthetic treatment covariates can be used when estimating the propensity score, while control observations and their synthetic counterparts provide a correction for generator-induced loss bias. The resulting propensity-score estimator can then be inserted into an augmented inverse probability weighting estimator \citep{rosenbaum1983central,robins1994estimation,glynn2010introduction}. We provide the construction in Section~B and a numerical illustration in Section~D of the Supplementary Material.


\section{Theoretical Properties}\label{sec_theory}
In this section, we first establish an upper bound for the difference between the minority and majority bias correction terms, $|\widehat\Delta_1 - \widehat\Delta_0|$. This result provides a theoretical guarantee for the construction of the bias-corrected loss function $L^{\rm bc}$ in (\ref{eq_bc_loss}). We then derive a lower bound for the minority bias correction term $\widehat\Delta_1$ of SMOTE, thereby illustrating that treating synthetic samples as real data may introduce bias. Finally, we show that the excess balanced risk of the bias-corrected predictor is controlled by the bias-transfer discrepancy and sampling fluctuations.

To establish finite-sample guarantees for the proposed bias-corrected loss and predictor, we introduce assumptions controlling the loss function, the complexity of the classifier class, the conditional sampling structure, and the transfer of synthetic loss bias across classes. Assumptions~\ref{asmp:bounded}--\ref{asmp:synthetic} are regularity conditions for concentration and empirical risk minimization. The substantive structural requirement is stated directly in Assumption~\ref{asmp:bias_transfer}, while a stronger transformation-based formulation is subsequently provided only as a sufficient condition.

We first impose a boundedness condition on the prediction function and the associated loss. This assumption ensures that the empirical risk and its bias correction terms are well-defined and uniformly controlled, which is essential for deriving concentration inequalities and uniform deviation bounds. Such boundedness conditions are standard in the analysis of classification methods with probabilistic outputs \citep{bousquet2002stability, bartlett2002rademacher}.
\begin{assumption}[Bounded prediction and loss]\label{asmp:bounded}
    There exists a constant $\varepsilon\in(0,1/2)$ such that for all $f\in\mathcal F$ and all $\bm x\in\mathcal X$, $f(\bm x)\in[\varepsilon,1-\varepsilon]$ and there exists a constant $M_\varepsilon>0$ such that $0 \le  \ell_f(\bm x,y)  \le M_\varepsilon$ for all $(\bm x,y)$. For instance, for the binary cross-entropy loss $\ell_f(\bm x,y)=-y\log f(\bm x)-(1-y)\log(1-f(\bm x))$, we have $M_\varepsilon=-\log\varepsilon$.
\end{assumption}

Next, we control the richness of the classifier class through its induced loss class. Bounding the Rademacher complexity of the normalized loss functions allows us to quantify the uniform generalization error of empirical risk minimization over $\mathcal F$. This assumption provides a standard measure of function class complexity and plays a central role in our finite-sample analysis.
\begin{assumption}[Function class complexity]\label{asmp:complexity}
    Let $\mathcal L=\{(\bm x,y)\mapsto \ell_f(\bm x,y)/M_\varepsilon:f\in\mathcal F\}$. For any probability distribution $Q$ on $\mathcal X\times\{0,1\}$ and sample size $m\geq1$, define
    \begin{align*}
        \mathfrak R_m(\mathcal L;Q) := \mathbb E_{\bm Z_{1:m}\sim Q^m,\,\bm\sigma}\left[\sup_{g\in\mathcal L}\left|\frac{1}{m}\sum_{i=1}^m \sigma_i\, g(\bm Z_i)\right|\right],
    \end{align*}
    where $\bm Z_1,\ldots,\bm Z_m$ are i.i.d. observations from $Q$, $\sigma_1,\ldots,\sigma_m$ are i.i.d. Rademacher random variables satisfying $\mathbb P(\sigma_i=1)=\mathbb P(\sigma_i=-1)=1/2$, and $(\sigma_i)_{i=1}^m$ are independent of $(\bm Z_i)_{i=1}^m$.
    There exists a deterministic constant $C_{\mathcal F}>0$ such that, for every $m\geq1$ and $y\in\{0,1\}$,
    \begin{align*}
        \mathfrak R_m(\mathcal L;\mathcal P_y\otimes\delta_y)
        &\leq \frac{C_{\mathcal F}}{\sqrt m},\\
        \mathfrak R_m\bigl(\mathcal L;\tilde{\mathcal P}_y(\mathcal G)\otimes\delta_y\bigr)
        &\leq \frac{C_{\mathcal F}}{\sqrt m}
        \quad\text{almost surely with respect to $\mathcal G$},
    \end{align*}
    where $\delta_y$ denotes the point mass at label $y$, and $\tilde{\mathcal P}_y(\mathcal G)$ denotes the conditional synthetic distribution given the generator $\mathcal G$ introduced in Assumption~\ref{asmp:synthetic}.
\end{assumption}

We now formalize the stochastic structure of the synthetic samples. Rather than assuming unconditional independence, we introduce a conditional independence framework in which synthetic data are generated from a random generator $\mathcal G$ and are independent of the evaluation samples given $\mathcal G$. This formulation accommodates many synthetic data generation mechanisms, including resampling and model-based procedures, while preserving tractable theoretical analysis.
\begin{assumption}[Synthetic conditional independence]\label{asmp:synthetic}
    There exists a random object $\mathcal G$ (the fitted synthetic data generator) such that, conditional on $\mathcal G$, the synthetic samples $\{\tilde{\bm X}_i^{(1)}\}_{i=1}^{\tilde n_1}$ are i.i.d. from $\tilde{\mathcal P}_1(\mathcal G)$ and $\{\tilde{\bm X}_i^{(0)}\}_{i=1}^{\tilde n_0}$ are i.i.d. from $\tilde{\mathcal P}_0(\mathcal G)$. Here $\mathcal G$ contains the generator-training data, the fitting randomness, and, if random, the synthetic sample sizes, but not the subsequent randomness used to draw the synthetic observations. Moreover, the majority correction samples $(\bm X_i)_{i\in\mathcal S_{0c}}$ are independent of $\mathcal G$ and of the synthetic draws.
\end{assumption}

\begingroup
We now state the structural condition used in the main theoretical results. Write $\Delta_y(f)$ to emphasize the dependence of the population loss biases in \eqref{eq_population_bias_1} and \eqref{eq_population_bias_0} on $f$.
\begin{assumption}[Uniform bias transfer]\label{asmp:bias_transfer}
    There exists a constant $\varepsilon_{\rm BT}\geq0$ such that, almost surely with respect to the fitted generator $\mathcal G$,
    \begin{align*}
        \sup_{f\in\mathcal F}\left|\Delta_1(f)-\Delta_0(f)\right|\leq\varepsilon_{\rm BT}.
    \end{align*}
\end{assumption}
Assumption~\ref{asmp:bias_transfer} requires the generator-induced loss discrepancies to be approximately transferable across the two classes. It does not require the class-specific population risks to be close, nor does it require either synthetic distribution to approximate its corresponding population distribution. The condition concerns only the difference between the two class-specific synthetic loss biases.

We next provide a transformation-based formulation as one interpretable sufficient condition for Assumption~\ref{asmp:bias_transfer}. The following conditions are not required by the main theoretical results.
\begin{assumption}[Transformation-based sufficient conditions]\label{asmp_transformation}
    Consider the support of covariates $\mathcal X \subseteq \mathbb R^d$. Suppose the following conditions are satisfied uniformly for all $f \in \mathcal F$:
    \begin{enumerate}
        \item[(A1)] \textbf{Distribution transformation.} There exists a measurable function $T:\mathcal X \to \mathcal X$ and a constant $\varepsilon_T$ such that 
        \begin{align*}
            \mathcal W_1(\mathcal P_1, (\mathcal P_0)_{\# T})\le \varepsilon_T, \quad \text{and }\quad \mathcal W_1(\tilde{\mathcal P}_1, (\tilde{\mathcal P}_0)_{\#T}) \leq \varepsilon_T\quad \text{a.s. conditional on } \mathcal G,
        \end{align*}
        where $(\cdot)_{\#T}$ denotes the pushforward distribution induced by transformation $T$, and $\mathcal W_1(\cdot, \cdot)$ represents the Wasserstein--1 distance. 
        \item[(A2)] \textbf{Lipschitz smoothness of the transformed loss.} There exists a constant $L_\ell > 0$ such that for all $\bm x_1, \bm x_2\in\mathcal X$ and $y\in\{0, 1\}$, 
        \begin{align*}
            |\ell_f(T(\bm x_1), y) - \ell_f(T(\bm x_2), y)| \leq& L_\ell\|\bm x_1 - \bm x_2\|_2,
            \\
            |\ell_f(\bm x_1, y) - \ell_f(\bm x_2, y)| \leq& L_\ell\|\bm x_1 - \bm x_2\|_2.
        \end{align*}
        \item[(A3)] \textbf{Transformation compatibility of the loss.}
There exists a constant $\varepsilon_h\geq 0$ such that, uniformly for all $f\in\mathcal F$,
\begin{align*}
\left|
\mathbb E_{(\mathcal P_0)_{\#T}}\ell_f(\bm X,1)
-
\mathbb E_{\mathcal P_0}\ell_f(\bm X,0)
\right|
&\leq \varepsilon_h,
\\
\left|
\mathbb E_{(\tilde{\mathcal P}_0)_{\#T}}\ell_f(\bm X,1)
-
\mathbb E_{\tilde{\mathcal P}_0}\ell_f(\bm X,0)
\right|
&\leq \varepsilon_h
\qquad \text{almost surely with respect to }\mathcal G.
\end{align*}
Here $\tilde{\mathcal P}_0$ denotes the conditional synthetic distribution given the fitted generator $\mathcal G$.
    \end{enumerate}
\end{assumption}

\begin{lemma}[Transformation-based bias transfer]\label{lem:transformation_bias_transfer}
    Under Assumption~\ref{asmp_transformation}, Assumption~\ref{asmp:bias_transfer} holds with
    \begin{align*}
        \varepsilon_{\rm BT}=2L_\ell\varepsilon_T+2\varepsilon_h.
    \end{align*}
\end{lemma}

Assumption~\ref{asmp_transformation} provides one transparent construction under which the generator-induced loss bias is transferable. It is stronger than Assumption~\ref{asmp:bias_transfer}: in particular, (A1) and (A3) additionally align the class-specific population risks. For this reason, it is used only as a sufficient condition and is not imposed in Proposition~\ref{prop_Delta0_minus_Delta1_upper_bound} or Theorem~\ref{thm_bc_ERM}. The transformation $T$ is an analytical device and does not need to be estimated or specified when implementing the proposed method.

We next give a toy example showing that bias transfer can hold even when neither synthetic distribution is close to its population counterpart. Consider
\begin{align*}
    \mathcal P_0&=\mathcal N(-\bm\mu,\bm\Sigma),
    &\mathcal P_1&=\mathcal N(\bm\mu,\bm\Sigma),
\end{align*}
where $\bm\mu\in\mathbb R^d$ is nonzero and $\bm\Sigma$ is positive definite. Let $T(\bm x)=-\bm x$. Suppose a reflection-equivariant synthetic procedure introduces the same shrinkage of the two class means toward the origin, yielding
\begin{align*}
    \widetilde{\mathcal P}_0&=\mathcal N(-\rho\bm\mu,\bm\Sigma),
    &\widetilde{\mathcal P}_1&=\mathcal N(\rho\bm\mu,\bm\Sigma),
\end{align*}
for some $0\leq\rho<1$. Then
\begin{align*}
    \mathcal P_1=(\mathcal P_0)_{\#T},
    \qquad
    \widetilde{\mathcal P}_1=(\widetilde{\mathcal P}_0)_{\#T},
\end{align*}
but, for $y\in\{0,1\}$,
\begin{align*}
    \mathcal W_1(\widetilde{\mathcal P}_y,\mathcal P_y)
    =(1-\rho)\|\bm\mu\|_2>0.
\end{align*}
Thus, the synthetic distributions retain a non-vanishing marginal distortion whenever $\rho<1$.

Consider the binary cross-entropy loss and the logistic prediction function $f_{\bm\beta}(\bm x)=\{1+\exp(-\bm\beta^\top\bm x)\}^{-1}$. Since $f_{\bm\beta}(-\bm x)=1-f_{\bm\beta}(\bm x)$,
\begin{align*}
    \ell_{f_{\bm\beta}}(T(\bm x),1)=\ell_{f_{\bm\beta}}(\bm x,0).
\end{align*}
Consequently,
\begin{align*}
    \Delta_1(f_{\bm\beta})
    &=\mathbb E_{\mathcal P_1}\ell_{f_{\bm\beta}}(\bm X,1)
      -\mathbb E_{\widetilde{\mathcal P}_1}\ell_{f_{\bm\beta}}(\bm X,1)
    \\
    &=\mathbb E_{\mathcal P_0}\ell_{f_{\bm\beta}}(\bm X,0)
      -\mathbb E_{\widetilde{\mathcal P}_0}\ell_{f_{\bm\beta}}(\bm X,0)
    =\Delta_0(f_{\bm\beta}).
\end{align*}
Hence, Assumption~\ref{asmp:bias_transfer} holds with $\varepsilon_{\rm BT}=0$ for this class, although both synthetic distributions remain systematically different from their targets. The example illustrates that the relevant requirement is preservation of the generator-induced distortion across classes rather than marginal synthetic-data fidelity. It is intended to illustrate the bias-transfer condition alone, rather than all regularity conditions of Theorem~\ref{thm_bc_ERM}.
\endgroup

\begin{proposition}
    \label{prop_Delta0_minus_Delta1_upper_bound}
    Suppose Assumptions~\ref{asmp:bounded}, \ref{asmp:synthetic}, and~\ref{asmp:bias_transfer} hold. Then for any fixed $f\in\mathcal F$ and any $\alpha\in(0,1)$, with probability at least $1-\alpha$,
    \begin{align*}
        |\widehat\Delta_1 - \widehat\Delta_0| \leq  \varepsilon_{\rm BT} + M_\varepsilon\sqrt{\frac{\log(8/\alpha)}{2}}\left(\frac{1}{\sqrt{n_1^*}}+\frac{1}{\sqrt{\tilde n_1}} + \frac{1}{\sqrt{n_{0c}}} + \frac{1}{\sqrt{\tilde n_0}}\right).
    \end{align*}
    Specifically, if there exist $0<c_1<c_2<1/2$ such that $c_1\le n_1/n\le c_2$ and $n_1^*,\tilde n_1,n_{0c},\tilde n_0\asymp n_0-n_1$, then for any $\alpha > 0$ and some constant $C<\infty$, with probability at least $1-\alpha$,
    \begin{align*}
        |\widehat\Delta_1 - \widehat\Delta_0| \leq  
        \varepsilon_{\rm BT} + C\sqrt{\frac{\log(8/\alpha)}{n_0-n_1}}.
    \end{align*}
\end{proposition}

Proposition~\ref{prop_Delta0_minus_Delta1_upper_bound} establishes an upper bound on the discrepancy between the empirical majority and minority correction terms. The structural component $\varepsilon_{\rm BT}$ directly measures the non-transferable part of the two class-specific generator-induced loss biases, while the remaining terms arise from finite-sample fluctuations. The result does not require either synthetic distribution to approximate its target marginal distribution closely. Under the stronger transformation-based sufficient conditions, Lemma~\ref{lem:transformation_bias_transfer} gives the interpretable choice $\varepsilon_{\rm BT}=2L_\ell\varepsilon_T+2\varepsilon_h$.

A key challenge with synthetic oversampling methods such as SMOTE is that the synthetic distribution $\tilde{\mathcal P}_1$ of the minority class may not match the true distribution $\mathcal P_1$. We characterize a first-order bias regime under a non-cancellation condition on the average loss drift induced by local interpolation. We then show that the bias-correction error can be controlled by the discrepancy between the minority and majority loss biases.

\begin{assumption}~
    \label{asmp_SMOTE_density}
    \begin{enumerate}
        \item[(B1)] The loss $\ell_f(\bm x, y)$ is Lipschitz in $\bm x$ with constant $L_\ell$ uniformly over $y \in \{0, 1\}$, i.e., $|\ell_f(\bm x_1, y) - \ell_f(\bm x_2, y)| \leq L_\ell \|\bm x_1 - \bm x_2\|_2$ for any $\bm x_1, \bm x_2\in \mathbb R^d$ and $y \in \{0, 1\}$. 
        \item[(B2)] $\mathcal P_1$ is supported on a bounded set $B(0,R)\subseteq\mathbb R^d$ and has a density $f_1$ satisfying $0<C_1\le f_1(\bm x)\le C_2<\infty$ for all $\bm x\in{\rm supp}(\mathcal P_1)$. In addition, there exist $r_0>0$ and $c_s\in(0,1]$ such that, for every $\bm x\in{\rm supp}(\mathcal P_1)$ and $0<r\le r_0$, the Lebesgue measure of $B(\bm x,r)\cap{\rm supp}(\mathcal P_1)$ is at least $c_s v_dr^d$, where $v_d$ is the volume of the unit ball in $\mathbb R^d$.
        \item[(B3)] 
        Let $\mathcal D_1=\{X_1,\ldots,X_{n_1}\}$. Conditional on $\mathcal D_1$, let $I$ be uniformly distributed over ${1,\ldots,n_1}$, let $J$ be uniformly sampled from the $K$ nearest neighbours of $X_I$, and let $U\sim\operatorname{Unif}(0,1)$ independently. There exist constants $C_3>0$, $s_f\in\{-1,1\}$, and a sequence $\eta_{n_1}\to0$ such that, with probability at least $1-\eta_{n_1}$ over $\mathcal D_1$,
        \begin{align*}
            s_f \mathbb E[\ell_f\{X_I + U(X_J - X_I), 1\} - \ell_f(X_I, 1) \mid \mathcal D_1] \ge C_3 \mathbb E[U\|X_J - X_I\|_2 \mid \mathcal D_1].
        \end{align*}
    \end{enumerate}
\end{assumption}

Assumption~\ref{asmp_SMOTE_density}~(B1) guarantees that the loss function $\ell_f$ is Lipschitz continuous with respect to $\bm x$ uniformly over $y\in\{0, 1\}$. (B2) ensures that the minority distribution has bounded support, a bounded density function on its support, and sufficient local support volume, including near the boundary.  (B3) requires that, after averaging over the random anchor point, its nearest neighbor, and the interpolation coefficient used by SMOTE, the synthetic point induces a non-vanishing change in the loss proportional to the average interpolation distance. Thus, individual interpolation directions may either increase or decrease the loss, provided that their aggregate effects do not completely cancel.  For a differentiable loss, this condition can be interpreted as requiring a non-vanishing average directional component of the loss gradient along the local interpolation directions selected by SMOTE. If this average first-order effect cancels, (B3) does not hold and the leading loss discrepancy may be of smaller order.

The following proposition shows that when the minority synthetic samples are generated by SMOTE, the induced bias cannot vanish too quickly. Specifically, under the average one-sided loss drift condition, the induced loss bias admits a lower bound of order $(K/n_1)^{1/d}$, corresponding to the typical scale of the local interpolation distances used by SMOTE.

\begin{proposition}
    \label{prop_Delta1_lower_bound}
    Suppose Assumptions~\ref{asmp:bounded} and~\ref{asmp_SMOTE_density} hold for a deterministic $f \in \mathcal F$, $d> 1$, $K$ is fixed, and conditional on $\mathcal D_1$, the synthetic minority samples are generated independently from $\tilde{\mathcal P}_1$ via SMOTE with parameter $K$. Then there exists a constant $c_L>0$, depending only on $(d,C_2,C_3,K)$, such that
    for any $\alpha\in(0, 1)$, with probability at least $ 1- \alpha - \eta_{n_1} - \eta_{K, n_1}$,
    \begin{align*}
        |\widehat\Delta_1| \geq c_L \left(\frac{K}{n_1}\right)^{1/d} - M_\varepsilon\sqrt{\frac{\log(6/\alpha)}{2n_1}} - M_\varepsilon\sqrt{\frac{\log(6/\alpha)}{2\tilde n_1}} - M_\varepsilon\sqrt{\frac{\log(6/\alpha)}{2n_1^*}},
    \end{align*}
    where $\eta_{K,n_1}\to0$ denotes the probability associated with the nearest-neighbour distance bound.
\end{proposition}
For fixed $K$, when $n_1^*$ and $\tilde n_1$ are of order $n_1$, the lower bound in Proposition~\ref{prop_Delta1_lower_bound} is asymptotically nontrivial for $d>2$, because its leading term decays more slowly than the sampling terms. For $d\le2$, the bound alone need not separate the bias from sampling fluctuations. In the nontrivial regime, the result shows that the synthetic-augmented loss function (\ref{eq_syn_loss}) need not serve as a close substitute for the balanced loss function (\ref{eq_bal_loss}), which may degrade the performance of the trained classifier. 

Beyond quantifying the bias induced by synthetic sampling, an important question is how such bias affects the learning procedure itself. The natural target is the population balanced risk, which is defined as 
\begin{align}
    \label{eq_population_balanced_risk}
    L^*(f) =  \frac{1}{2} \mathbb E_{\mathcal P_1}[\ell_f(\bm X, 1)] + \frac{1}{2} \mathbb E_{\mathcal P_0}[\ell_f(\bm X, 0)].
\end{align}
The population balanced risk function $L^*$ assigns equal weight to the two class-conditional risks, irrespective of their marginal class probabilities. 
Denote the corresponding population balanced risk minimizer as \begin{align}
    \label{eq_population_risk_minimizer}
    f^* = \underset{f\in\mathcal F}{\arg\min} \:L^*(f),
\end{align}
where $\mathcal F$ is the prediction-function class introduced in Section~\ref{sec:method}.
In practice, however, one does not have access to the population balanced loss $L^*$, but instead minimizes an empirical loss. With synthetic samples generated and the bias correction procedure,
we have the empirical minimizer $\widehat f^{\rm bc}$ of the bias-corrected loss function \eqref{eq_f_bc}.
Next, we investigate the following question: how close is the empirical bias-corrected minimizer $\widehat f^{\rm bc}$ to the population balanced minimizer $f^*$?

To answer this question, we use the uniform bias-transfer condition in Assumption~\ref{asmp:bias_transfer}. It directly requires the gap between the minority and majority generator-induced loss biases to remain controlled over $\mathcal F$. If $\varepsilon_{\rm BT}$ is small, correcting the minority synthetic loss using information from the majority group is reliable.

The following theorem then provides an upper bound on the excess population risk of the bias-corrected minimizer.
\begin{theorem}
    \label{thm_bc_ERM}
    Conditional on the realized class labels, suppose Assumptions~\ref{asmp:bounded}, \ref{asmp:complexity}, \ref{asmp:synthetic}, and~\ref{asmp:bias_transfer} hold, $\tilde n_1=n_0-n_1$, and there exist constants $0<c_1<c_2<1/2$ such that $c_1\le n_1/n\le c_2$. Then for any $\alpha \in (0, 1)$, there exists a constant $C$ depending only on $M_\varepsilon,c_1,c_2$ such that, with conditional probability at least $1-\alpha$,
    \begin{align*}
        L^*(\widehat f^{\rm bc}) - L^*(f^*) \leq & \frac{n_0-n_1}{n_0}\varepsilon_{\rm BT} \\
        &+ C(2C_{\mathcal F}+\sqrt{\log(12/\alpha)})\cdot\left\{\frac{1}{\sqrt{n_1}} + \frac{1}{\sqrt{n_0}} + \frac{1}{\sqrt{n_{0c}}} + \frac{1}{\sqrt{\tilde n_1}} + \frac{1}{\sqrt{\tilde n_0}}\right\}.
    \end{align*}
\end{theorem}
The bound in Theorem~\ref{thm_bc_ERM} shows that the excess population risk of the bias-corrected estimator is controlled by two types of terms:
\begin{enumerate}
    \item[(i)] A sample-proportion-weighted bias transfer term, $(n_0-n_1)\varepsilon_{\rm BT}/n_0$, which measures the worst-case mismatch of the bias correction from the majority to the minority group.
    \item[(ii)] Complexity terms of the order of inverse square root of sample sizes, which quantify the statistical fluctuations from randomness of finite samples in each component dataset.
\end{enumerate}
Thus, under the structural bias-transfer condition, Theorem~\ref{thm_bc_ERM} shows that the bias-corrected empirical risk minimizer achieves population risk close to that of the optimal $f^*$, up to statistical and transfer errors. Under i.i.d. Bernoulli sampling, $(n_0-n_1)/n_0$ converges in probability to $(\pi_0-\pi_1)/\pi_0$.

To compare synthetic augmentation and bias correction directly, consider the population counterparts $\overline L^{\rm syn}$ and $\overline L^{\rm bc}$ obtained by replacing the empirical averages in $L^{\rm syn}$ and $L^{\rm bc}$ by their corresponding expectations. When $\tilde n_1=n_0-n_1$, let $w_n=(n_0-n_1)/(2n_0)$ and write $\Delta_y(f)$ to emphasize the dependence of the loss bias on $f$. A direct calculation gives
\begin{align*}
    \overline L^{\rm syn}(f)-L^*(f) &= -w_n\Delta_1(f),\\
    \overline L^{\rm bc}(f)-L^*(f) &= w_n\{\Delta_0(f)-\Delta_1(f)\}.
\end{align*}
Thus, synthetic augmentation is governed by the absolute minority loss bias, whereas bias correction is governed by the component that cannot be transferred from the majority class. Bias correction is most beneficial when the reduction in this structural discrepancy exceeds the additional sampling error required to estimate $\Delta_0(f)$.


\section{Data Analysis}\label{sec:data}

We evaluated the proposed bias-correction procedure on the EchoNet-Dynamic dataset \citep{echonetdataset,ouyang2020video} using a binary classification task for identifying severe left ventricular dysfunction. EchoNet-Dynamic consists of apical four-chamber echocardiographic videos labeled by left ventricular ejection fraction (LVEF). For this analysis, we extracted the first frame from each video included in the analysis and categorized the resulting images into three groups according to LVEF: normal (LVEF $\geq 55\%$, $n=6{,}961$), mild dysfunction ($40\% \leq$ LVEF $<55\%$, $n=805$), and severe dysfunction (LVEF $<40\%$, $n=1{,}246$). We then formulated a binary classification problem in which severe dysfunction constituted the minority class and normal and mild dysfunction were combined to form the majority class. For each experimental replicate, the images were randomly divided into training, validation, and test sets in proportions of $60\%$, $20\%$, and $20\%$, respectively, using stratified sampling.

For each data split, we trained a ResNet-18 convolutional neural network \citep{he2016deep} using only the real training images, with the validation set used for model selection. ResNet-18 consists of four stages of residual blocks with channel dimensions $64$, $128$, $256$, and $512$, where identity shortcut connections facilitate the learning of residual mappings. The network was randomly initialized and trained to classify severe versus non-severe left ventricular dysfunction. After model selection, the network was frozen and the output of its global average pooling layer was used as a $512$-dimensional image representation. All subsequent synthetic-data generation, bias correction, and classification were performed in this fixed feature space, while the test set was reserved exclusively for final evaluation.

We considered binary cross-entropy (BCE) loss $\ell_f(\bm x,y) = -y\log f(\bm x)-(1-y)\log\{1-f(\bm x)\}$.
Consistent with the balanced population risk studied in our theoretical analysis, performance was evaluated using balanced test BCE,
\[
    \widehat L_{\mathrm{bal}} = \frac{1}{2}\widehat L_{1,\mathrm{test}} + \frac{1}{2}\widehat L_{0,\mathrm{test}},
\text{ where }
    \widehat L_{k,\mathrm{test}} = \frac{1}{n_{k,\mathrm{test}}} \sum_{i:Y_i=k} \ell_f(\bm X_i,k) \text{ for } k\in\{0,1\}.
\]
Thus, the two classes contribute equally to the evaluation criterion regardless of their prevalence in the test data.

To evaluate bias correction under controlled synthetic distribution shift, we constructed a centroid-shrinkage version of SMOTE. Let $\widehat{\bm \mu}_k$ denote the empirical feature centroid of class $k$ in the training data. For an ordinary SMOTE sample $X_{\mathrm{SMOTE}}^{(k)}$, we generated
$$
    \widetilde{\bm X}^{(k)}(\rho) = \widehat{\bm \mu}_k + \rho \left\{ \bm X_{\mathrm{SMOTE}}^{(k)} - \widehat{\bm \mu}_k \right\}, \qquad k\in\{0,1\},
$$
where $\rho\in\{1.0,0.8,0.6,0.4,0.2\}$. The case $\rho=1$ corresponds to ordinary SMOTE, while smaller values of $\rho$ increasingly concentrate synthetic observations toward their respective class centroids. The same transformation was applied to both classes. For each $\rho$, we compared synthetic augmentation with its bias-corrected counterpart using the procedure described in Section~\ref{sec:method}.

We repeated the complete analysis over $20$ prespecified random seeds, with an independent stratified training/validation/test split and ResNet-18 representation for each replicate. We define the improvement due to bias correction as $ \Delta_{\mathrm{BCE}}(\rho) = \widehat L_{\mathrm{bal}}^{\mathrm{Syn}}(\rho) - \widehat L_{\mathrm{bal}}^{\mathrm{BC}}(\rho)$, such that a positive value favors bias correction.
\begin{table}[t]
\centering
\setlength{\tabcolsep}{5.5pt}
\caption{Balanced test BCE for synthetic augmentation with and without bias correction over $20$ replicates. Results are mean (SD). Improvement is defined as $\mathrm{BCE}_{\mathrm{Syn}}-\mathrm{BCE}_{\mathrm{BC}}$.}
\label{tab:echonet-centroid}
\begin{tabular}{cccccc}
\toprule
$\rho$
& Synthetic
& Bias Correction
& Improvement
& 95\% CI
& BC Wins \\
\midrule
1.0
& $0.5705\;(0.0167)$
& $0.5724\;(0.0207)$
& $-0.0019$
& $[-0.0046,\;0.0009]$
& $7/20$ \\
0.8
& $0.5759\;(0.0169)$
& $0.5662\;(0.0157)$
& $\mathbf{0.0097}$
& $[0.0083,\;0.0111]$
& $20/20$ \\
0.6
& $0.5798\;(0.0165)$
& $0.5638\;(0.0142)$
& $\mathbf{0.0160}$
& $[0.0142,\;0.0177]$
& $20/20$ \\
0.4
& $0.5815\;(0.0162)$
& $0.5630\;(0.0137)$
& $\mathbf{0.0185}$
& $[0.0165,\;0.0205]$
& $20/20$ \\
0.2
& $0.5820\;(0.0160)$
& $0.5629\;(0.0136)$
& $\mathbf{0.0191}$
& $[0.0169,\;0.0213]$
& $20/20$ \\
\bottomrule
\end{tabular}
\end{table}
As shown in Table~\ref{tab:echonet-centroid}, bias correction provided little benefit under ordinary SMOTE ($\rho=1$), with a mean paired difference of $-0.0019$ (95\% CI: $-0.0046$ to $0.0009$). In contrast, once centroid shrinkage was introduced, bias correction consistently reduced balanced BCE. The mean improvement increased from $0.0097$ at $\rho=0.8$ to $0.0191$ at $\rho=0.2$, and bias correction outperformed the corresponding uncorrected synthetic procedure in all $20$ replicates for every $\rho<1$. Moreover, stronger shrinkage was associated with a larger improvement, with a mean within-replicate slope of $0.02534$ (95\% CI: $0.02038$ to $0.03030$; $p=1.77\times10^{-9}$) with respect to shrinkage strength $1-\rho$.

The increasing benefit of bias correction was accompanied by increasing estimated synthetic loss bias. The mean $\widehat\Delta_0$ increased from approximately $-0.0025$ at $\rho=1$ to $0.0937$ at $\rho=0.2$, and $|\widehat\Delta_0|$ was strongly associated with the balanced-BCE improvement across the $100$ seed-by-shrinkage combinations (Spearman correlation $0.737$, $p=2.45\times10^{-18}$). These results indicate that bias correction has little effect when synthetic bias is small, but increasingly mitigates the excess balanced risk as the synthetic distribution becomes more distorted. In this experiment, the correction mitigates the component of synthetic-data distortion that is transferable across classes.

Additional simulation results and an analysis of the MNIST handwritten digit dataset \citep{lecun1998mnist,mnist554dataset} are provided in Section~D of the Supplementary Material.

\section{Discussion}
We propose a novel bias correction framework for imbalanced classification with synthetic data augmentation, motivated by the observation that treating synthetic data as real data often introduces systematic bias due to imperfect data generation. The proposed method estimates and corrects the bias by leveraging majority-class data, where the majority samples are partitioned into generation and correction subsets, enabling bias estimation and correction through discrepancies between synthetic and real majority samples. 
Theoretical results establish that the correction error is governed by the bias-transfer discrepancy and sampling fluctuations rather than by the marginal fidelity of either synthetic distribution alone. Empirical studies, including simulations and analyses of the MNIST and EchoNet-Dynamic datasets, show that the correction is most useful when the synthetic generator introduces appreciable distortion and offers limited gains when that distortion is small.

This approach has several limitations: (i) The current theory characterizes bias through a scalar loss function, which may inadequately capture complex discrepancies in high-dimensional settings. (ii) When synthetic generators already produce highly accurate samples, the additional gains from bias correction are limited and may not justify the computational cost. (iii) While empirical results suggest that the method extends naturally to multi-class problems, a formal theoretical justification is currently lacking.
Future work will focus on developing richer discrepancy measures beyond loss-based bias, establishing principled criteria to identify scenarios where bias correction is beneficial, and providing a rigorous theoretical extension to multi-class classification.

\section*{Acknowledgements}

P. L. and A. R. Z. were partially supported by NIH Grant R01HL169347; Z. M. and A. R. Z. were partially supported by NIH Grant R01HL168940; A. R. Z. was also partially supported by NSF Grant CAREER 2203741. L. Z. was partially supported by NSF Grant CAREER 2340241 and Renaissance Philanthropy ``AI for Math'' Fund.

\section*{Data Availability Statement}
The MNIST data used in this study are publicly available through OpenML \citep[dataset 554; \url{https://www.openml.org/d/554};][]{mnist554dataset}. The EchoNet-Dynamic data are available from the dataset custodians \citep[\url{https://echonet.github.io/dynamic/};][]{echonetdataset} subject to their access terms. Code and additional reproducibility materials are available from the corresponding author upon request.

\bibliography{ref}
\bibliographystyle{apalike}

\newpage
\appendix

We provide methodological extensions to multi-task learning in Section~\ref{suppsec_MTL}
and average treatment effect estimation in Section~\ref{suppsec_ATE}, followed by proofs of the main-paper theoretical results in Section~\ref{suppsec_proof}. Additional numerical studies are provided in Section~\ref{sec_simu}. In Section~\ref{suppsec_syn}, we review commonly used synthetic-data generators.
\begin{sloppypar}

\section{Extension to Multi-Task Learning}\label{suppsec_MTL}
\subsection{Methodology}\label{sec:within-source}

In this subsection, we focus on applying bias correction techniques to datasets involving multiple related tasks, a scenario commonly addressed by multi-task learning (MTL). MTL is a machine learning paradigm where multiple related tasks are learned simultaneously, enabling the model to leverage shared information and learn a common robust representation \citep{caruana1997multitask, zhang2021survey}. For example, in genomic studies, researchers analyze gene expression data from different regional populations to identify genetic markers for specific diseases such as Alzheimer's disease \citep{zhang2011multi}. In this case, each regional population is regarded as a separate learning task. While the goal of identifying Alzheimer's disease is shared, the genetic and environmental differences between populations lead to unique data distributions. This makes it necessary to utilize a multi-task learning framework to leverage the common structure. However, the number of individuals with the disease is typically smaller than the number of healthy individuals, creating a within-task imbalance problem \citep{wu2018imbalanced, guo2025revisit}. We apply the bias correction procedure to such imbalanced MTL problems to improve the predictive performance by leveraging information from all tasks. 

Consider datasets from $K$ learning tasks and for each task $k = 1,\ldots,K$, there are $n_k$ samples of covariates $\bm X_{ki}\in\mathbb R^d$ and class labels $Y_{ki} \in \{0, 1\}$ for $i = 1, \ldots, n_k$. 
Under the imbalanced setting, the class labels are imbalanced within each task such that the marginal probability $\pi_{k1} = \mathbb P(Y_{ki}=1) < 1/2$. The dependence structure of the class labels on the covariates is captured by the following Bernoulli model:
\begin{align}
    \label{eq_bernoulli_model}
    \mathbb P(Y_{ki}=1\mid \bm X_{ki}=\bm x) = \sigma(\bm x^\top \bm B \bm\alpha_{k})\quad  \text{ for } i = 1, \ldots, n_k,
\end{align}
where $\sigma(t) = 1/(1+\exp(-t))$ denotes the logistic function, $\bm B\in\mathbb R^{d\times r}$ denotes the shared coefficient matrix across $K$ tasks and $\bm \alpha_1, \ldots, \bm \alpha_K \in\mathbb R^r$ are task-specific latent loading vectors. Denote $\bm M = \bm B(\bm \alpha_1, \ldots, \bm\alpha_K) \in\mathbb R^{d\times K}$ as the unknown coefficient matrix with ${\rm rank}(\bm M) = r$. Consider the task-specific coefficient vector $\bm \beta_k = \bm B\bm\alpha_k\in\mathbb R^d$ as the $k$th column of $\bm M$ for $k = 1,\ldots,K$. Our goal is to learn the left singular vector space of the shared matrix $\bm B$.

For any $\bm\beta\in\mathbb R^d$, the prediction function is provided by $f(\bm x) = \sigma(\bm x^\top \bm\beta)$ and denote the loss function as 
\begin{align*}
    \ell_f(\bm x, y) = \ell(\bm x, y; \bm\beta) = -y\log(\sigma(\bm x^\top \bm\beta)) - (1-y)\log(1-\sigma(\bm x^\top \bm\beta)).
\end{align*}
For each task, we obtain the estimator $\widehat{\bm\beta}_k^{\rm raw}$ from $(\bm X_{ki}, Y_{ki})_{i=1}^{n_k}$ by minimizing the loss function from the raw data $L^{\rm raw}(\bm\beta) =\frac{1}{n_k}\sum_{i=1}^{n_k} \ell(\bm X_{ki}, Y_{ki}; \bm\beta).$ Consider the minority synthetic samples $(\tilde{\bm X}_{ki}^{(1)})_{i=1}^{\tilde n_{k1}}$ and majority synthetic samples $(\tilde{\bm X}_{ki}^{(0)})_{i=1}^{\tilde n_{k0}}$. We can also obtain synthetic and bias-corrected loss functions by 
\begin{align*}
    L^{\rm syn}(\bm\beta) =& \frac{1}{n_k + \tilde n_{k1}}\left[\sum_{i=1}^{n_k}\ell(\bm X_{ki}, Y_{ki}; \bm\beta) + \sum_{i=1}^{\tilde n_{k1}} \ell(\tilde{\bm X}_{ki}^{(1)}, 1; \bm\beta)\right],
    \\
    L^{\rm bc}(\bm\beta) =& \frac{1}{n_k + \tilde n_{k1}}\left[\sum_{i=1}^{n_k}\ell(\bm X_{ki}, Y_{ki}; \bm\beta) + \tilde n_{k1}\cdot\left\{\frac{1}{\tilde n_{k1}}\sum_{i=1}^{\tilde n_{k1}} \ell(\tilde{\bm X}_{ki}^{(1)}, 1; \bm\beta) + \widehat\Delta_{k0}\right\}\right],
\end{align*}
where $\widehat\Delta_{k0}$ denotes the bias correction term from the majority samples in the correction set $\mathcal S_{kc}$ and majority synthetic samples:
\begin{align*}
    \widehat\Delta_{k0} =& \frac{1}{|\mathcal S_{kc}|}\sum_{i\in\mathcal S_{kc}}\ell(\bm X_{ki}, 0; \bm\beta) - \frac{1}{\tilde n_{k0}}\sum_{i=1}^{\tilde n_{k0}} \ell(\tilde{\bm X}_{ki}^{(0)}, 0; \bm\beta).
\end{align*}
Denote $\widehat{\bm \beta}_k^{\rm raw}$, $\widehat{\bm \beta}_k^{\rm syn}$ and $\widehat{\bm \beta}_k^{\rm bc}$ as the minimizers of the raw loss $L^{\rm raw}$,  synthetic-augmented loss $L^{\rm syn}$ and bias-corrected loss $L^{\rm bc}$, respectively. Throughout this paper, we assume that each minimum is attained. If any objective admits multiple minimizers, the corresponding estimator is defined as an arbitrary measurable selection from its argmin set. 

Next, we consider the estimation of the left singular matrix $\bm U$ from any coefficient estimators $(\widehat{\bm\beta}_k)_{k=1}^K$. First, collect all estimators into $\widehat{\bm M} = (\widehat{\bm \beta}_1, \dots, \widehat{\bm\beta}_K) \in\mathbb R^{d\times K}$. Next, conduct eigendecomposition of $\widehat{\bm M} \widehat{\bm M}^T$ such that $\widehat{\bm M} \widehat{\bm M}^T = \widehat{\bm U}'\widehat{\bm\Lambda}\widehat{\bm U}'^{\top}$, where $\widehat{\bm U}'\in\mathbb R^{d\times d}$ is an orthonormal eigenvector matrix satisfying $(\widehat{\bm U'})^{\top} \widehat{\bm U}' = \bm I_d$ and $\widehat{\bm\Lambda} = {\rm diag}(\widehat\lambda_1, \dots, \widehat\lambda_d)$ is a diagonal eigenvalue matrix with decreasing eigenvalues $\widehat\lambda_1 \geq \cdots \geq \widehat\lambda_d$. The rank of the latent embedding matrix is estimated using an eigenvalue ratio criterion that has been widely adopted for determining the number of latent factors in high-dimensional spectral and factor models \citep{ahn2013eigenvalue,fan2019farmtest,wang2026statistical}. Specifically, the rank is selected by maximizing the eigenvalue ratio such that $\widehat r = \arg\max_{1\leq r \leq d_-} \widehat\lambda_{r}/\widehat\lambda_{r+1}$, where $d_- < d$ is a constant to avoid numerical instability caused by extremely small eigenvalues. Finally, take the first $\widehat r$ columns of $\widehat{\bm U}'$ to obtain the estimated shared embedding matrix $\widehat{\bm U} = \widehat{\bm U}'_{1:\widehat r}$. Applying the same construction to different coefficient estimators yields the corresponding latent embedding matrix estimators. In particular, using the raw estimator $\widehat{\bm\beta}_k^{\rm raw}$, the synthetic estimator $\widehat{\bm\beta}_k^{\rm syn}$, and the bias-corrected estimator $\widehat{\bm\beta}_k^{\rm bc}$ in place of $\widehat{\bm\beta}_k$ results in $\widehat{\bm U}^{\rm raw}$, $\widehat{\bm U}^{\rm syn}$, and $\widehat{\bm U}^{\rm bc}$, respectively.

Suppose we are then provided with samples from a new task $(\bm X_{K+1, i}, Y_{K+1, i})_{i=1}^{n_{K+1}}$ from the following Bernoulli model with the same shared structure:
\begin{align*}
    \mathbb P(Y_{K+1, i} = 1 \mid \bm X_{K+1, i} = \bm x) = \sigma(\bm x^\top \bm B\bm\alpha_{K+1}).
\end{align*}
The estimated shared embedding matrix $\widehat{\bm U}$ helps us to estimate the coefficient in a lower dimension $\widehat r$ rather than $d$. To obtain the estimator, we can first project the covariates into a lower-dimensional embedding subspace by $\widehat{\bm Z}_{K+1, i} = \widehat{\bm U}^\top \bm X_{K+1,i}$ for each $i = 1, \dots, n_{K+1}$. Next, obtain $\widehat{\bm\theta}_{K+1}$ which minimizes the loss function, for example, $L^{\rm raw}(\bm\theta)$ on the dataset $(\widehat{\bm Z}_{K+1, i}, Y_{K+1, i})_{i=1}^{n_{K+1}}$. Note that the empirical loss function $L^{\rm raw}$ can be replaced by $L^{\rm syn}$ and $L^{\rm bc}$ depending on the imbalance of task $K+1$. Then project $\widehat{\bm\theta}_{K+1}$ back to the coefficient space by $\widehat{\bm\beta}_{K+1} = \widehat{\bm U}\widehat{\bm\theta}_{K+1}$ and obtain the prediction function $\widehat{f}_{K+1}(\bm x) = \sigma(\bm x^\top \widehat{\bm\beta}_{K+1})$. By substituting $\widehat{\bm U}$ with $\widehat{\bm U}^{\rm raw}$, $\widehat{\bm U}^{\rm syn}$, or $\widehat{\bm U}^{\rm bc}$, we obtain the corresponding coefficient estimator and prediction function for the new task.



\section{Extension to Average Treatment Effect Estimation}\label{suppsec_ATE}
\subsection{Methodology}
Our proposed methodology has an application to average treatment effect (ATE) estimation, one fundamental problem in causal inference \citep{rubin1974estimating,lunceford2004stratification}. In this context, imbalanced data often arises when the number of individuals receiving the treatment is significantly smaller than the number of individuals receiving the control. This imbalance makes the estimation of the ATE, more specifically, the expected outcome for the minority group, less reliable due to the limited sample size. This directly degrades the credibility and robustness of the final ATE estimate. Consequently, addressing this imbalanced data is essential for accurate causal inference.

Suppose $Y(1)$ and $Y(0)$ are the potential responses under treatment $Z=1$ and control $Z=0$. 
The observed response is a function of the potential responses and the treatment indicator: $Y = ZY(1) + (1-Z) Y(0)$. Then the ATE is defined as $\tau = \mathbb E\{Y(1)\} - \mathbb E\{Y(0)\}$.

Consider i.i.d. observations $(\bm X_i, Y_i, Z_i)_{i=1}^n$, where $Y_i = Z_i Y_i(1) + (1-Z_i)Y_i(0)$ for $i = 1,\ldots,n$,
$\bm X_i\in\mathbb{R}^d$ is a vector of covariates, and $Z_i\in \{0, 1\}$ is the treatment indicator. Suppose there are $n_1$ and $n_0$ samples from the treatment group and control group, respectively, and define the treated and control covariate indices as $\mathcal S_1 = \{i: Z_i = 1\}$ and $\mathcal S_0 = \{i: Z_i = 0\}$. In the imbalanced case where $n_1 < n_0$, we aim to estimate the ATE augmented with synthetic data by the bias correction approach.

Consider the propensity score \citep{rosenbaum1983central}, which is defined as the conditional probability of a sample receiving treatment given the corresponding covariate $\bm X_i = \bm x$:
$e^*(\bm x) = \mathbb P(Z_i = 1\mid \bm X_i = \bm x).$
For the propensity score, suppose we have an estimating model denoted as $e(\bm x)$. Similarly, for the conditional means of the responses given the covariates $\mu_1^*(\bm x) = \mathbb E\{Y_i(1)\mid \bm X_i = \bm x\}$ under treatment and $\mu_0^*(\bm x) = \mathbb E\{Y_i(0)\mid \bm X_i = \bm x\}$ under control, suppose we have the estimating models $\mu_1(\bm x)$ and $\mu_0(\bm x)$, respectively. We consider the augmented inverse propensity weighting (AIPW) estimators \citep{robins1994estimation, glynn2010introduction}:
\begin{align}
    \label{eq_AIPW_mu1}
    \widehat{\mu}_1^{\rm AIPW} =& \frac{1}{n} \sum_{i=1}^n \left[\frac{Z_i \{Y_i - \mu_1(\bm X_i)\}}{e(\bm X_i)} + \mu_1(\bm X_i)\right],
    \\
    \label{eq_AIPW_mu0}
    \widehat{\mu}_0^{\rm AIPW} =& \frac{1}{n} \sum_{i=1}^n \left[\frac{(1-Z_i) \{Y_i - \mu_0(\bm X_i)\}}{1-e(\bm X_i)} + \mu_0(\bm X_i)\right],
    \\
    \label{eq_AIPW_ATE}
    \widehat\tau^{\rm AIPW} =& \widehat\mu_1^{\rm AIPW} - \widehat\mu_0^{\rm AIPW}.
\end{align}

With the observations $(\bm X_i, Y_i, Z_i)_{i=1}^n$, we can first fit separate regression models of the responses on the covariates and obtain the coefficient estimators $\widehat{\bm\beta}_1$ and $\widehat{\bm\beta}_0$ under treatment and control, respectively. Let $\widehat{\mu}_1(\bm x) = \bm x^\top\widehat{\bm\beta}_1$ and $\widehat{\mu}_0(\bm x) = \bm x^\top\widehat{\bm\beta}_0$ be the estimated responses under treatment and control, respectively. 

Next, we consider propensity-score estimation. Suppose we are interested in a loss function $\ell_f$, e.g., logistic loss, for a prediction function $f: \mathbb R^d\to (0,1)$. With the raw data, we obtain the propensity-score estimator $\widehat f^{\rm raw}$ by minimizing the empirical loss function $L^{\rm raw}$ in (\ref{eq_raw_loss}). If there are $\tilde n_1$ synthetic minority covariate samples $\tilde{\bm X}_i^{(1)}$, we obtain the synthetic-augmented estimator $\widehat f^{\rm syn}$ by minimizing $L^{\rm syn}$ in (\ref{eq_syn_loss}). Partition the control index set into a generation set $\mathcal S_{0g}$ and a correction set $\mathcal S_{0c}$. After applying the same synthetic generator to obtain $\tilde n_0$ synthetic control covariates $\tilde{\bm X}_i^{(0)}$ from the generation set, we obtain the bias-corrected estimator $\widehat f^{\rm bc}$ by minimizing $L^{\rm bc}$ in (\ref{eq_bc_loss}).

Plugging in the treatment and control models $\widehat{\mu}_1(\bm x)$ and $\widehat{\mu}_0(\bm x)$ as well as the propensity-score estimate $\widehat e(\bm x) \in \{\widehat f^{\rm raw}(\bm x), \widehat f^{\rm syn}(\bm x), \widehat f^{\rm bc}(\bm x)\}$ into (\ref{eq_AIPW_ATE}) yields the corresponding estimators $\widehat{\tau}^{\rm raw}$, $\widehat\tau^{\rm syn}$, and $\widehat\tau^{\rm bc}$. To avoid reusing the same observations for nuisance estimation and AIPW evaluation, the outcome regressions and propensity score should be estimated by sample splitting or cross-fitting. The estimated propensity score should also be clipped away from zero and one before it is inserted into the AIPW score. Section~\ref{suppsec_ATE_simulation} provides a numerical illustration of the three implementations.


\section{Proofs of Main Results}\label{suppsec_proof}
\subsection{Proof of Lemma~\ref{lem:transformation_bias_transfer}}
\begin{proof}
For any $f\in\mathcal F$, let $h_f(\bm x)=\ell_f(T(\bm x),1)-\ell_f(\bm x,0)$. Adding and subtracting the transformed majority risks gives
\begin{align*}
\Delta_1(f)-\Delta_0(f)
=&\,\mathbb E_{\mathcal P_0}h_f(\bm X)-\mathbb E_{\widetilde{\mathcal P}_0}h_f(\widetilde{\bm X})
\\
&+\mathbb E_{\mathcal P_1}\ell_f(\bm X,1)-\mathbb E_{(\mathcal P_0)_{\#T}}\ell_f(\bm X,1)
\\
&+\mathbb E_{(\widetilde{\mathcal P}_0)_{\#T}}\ell_f(\widetilde{\bm X},1)-\mathbb E_{\widetilde{\mathcal P}_1}\ell_f(\widetilde{\bm X},1).
\end{align*}
By Assumption~\ref{asmp_transformation}~(A1)--(A2) and the Kantorovich--Rubinstein duality, the absolute values of the last two lines are each bounded by $L_\ell\varepsilon_T$. Assumption~\ref{asmp_transformation}~(A3) gives
\begin{align*}
\left|\mathbb E_{\mathcal P_0}h_f(\bm X)-\mathbb E_{\widetilde{\mathcal P}_0}h_f(\widetilde{\bm X})\right|\leq2\varepsilon_h.
\end{align*}
These bounds hold uniformly over $f\in\mathcal F$ and almost surely with respect to $\mathcal G$. Hence
\begin{align*}
\sup_{f\in\mathcal F}|\Delta_1(f)-\Delta_0(f)|\leq2L_\ell\varepsilon_T+2\varepsilon_h,
\end{align*}
which proves the result.
\end{proof}

\subsection{Proof of Proposition~\ref{prop_Delta0_minus_Delta1_upper_bound}}
\label{supp_subsec_Delta0_minus_Delta1_upper_bound}

\begin{proof}
The population quantities involving the synthetic distributions are interpreted conditional on the fitted generator $\mathcal G$. By the triangle inequality, we have
\begin{align}
\label{supeq_Delta_decomposition}
|\widehat\Delta_1 - \widehat\Delta_0|
\leq
|\Delta_1 - \Delta_0|
+
|\widehat\Delta_1 - \Delta_1|
+
|\widehat\Delta_0 - \Delta_0|,
\end{align}
where
\begin{align*}
\Delta_1
&=
\mathbb E_{\mathcal P_1}\ell_f(\bm X,1)
-
\mathbb E_{\tilde{\mathcal P}_1}\ell_f(\tilde{\bm X},1),\quad 
\Delta_0
=
\mathbb E_{\mathcal P_0}\ell_f(\bm X,0)
-
\mathbb E_{\tilde{\mathcal P}_0}\ell_f(\tilde{\bm X},0).
\end{align*}

By Assumption~\ref{asmp:bias_transfer},
\begin{align}
\label{supeq_pop_Delta_bound}
|\Delta_1-\Delta_0|\leq\varepsilon_{\rm BT}
\end{align}
almost surely with respect to $\mathcal G$.

We next control the empirical deviations.
Since $\ell_f\in[0,M_\varepsilon]$ by
Assumption~\ref{asmp:bounded}, and
$\bm X_1^*,\ldots,\bm X_{n_1^*}^*$ are i.i.d. from
$\mathcal P_1$, Hoeffding's inequality gives, for any $t>0$,
\begin{align}
\label{supeq_Delta1_real_term}
\mathbb P\left(
\left|
\frac{1}{n_1^*}
\sum_{i=1}^{n_1^*}
\ell_f(\bm X_i^*,1)
-
\mathbb E_{\mathcal P_1}\ell_f(\bm X,1)
\right|>t
\right)
\leq
2\exp\left\{
-\frac{2n_1^*t^2}{M_\varepsilon^2}
\right\}.
\end{align}

For the synthetic minority observations, conditional on $\mathcal G$,
Assumption~\ref{asmp:synthetic} implies that
$\tilde{\bm X}_1^{(1)},\ldots,
\tilde{\bm X}_{\tilde n_1}^{(1)}$
are i.i.d. from $\tilde{\mathcal P}_1$. Hence
\begin{align*}
\mathbb P\left(
\left.
\left|
\frac{1}{\tilde n_1}
\sum_{i=1}^{\tilde n_1}
\ell_f(\tilde{\bm X}_i^{(1)},1)
-
\mathbb E_{\tilde{\mathcal P}_1}
\ell_f(\tilde{\bm X},1)
\right|>t
\,\right|\mathcal G
\right)
\leq
2\exp\left\{
-\frac{2\tilde n_1t^2}{M_\varepsilon^2}
\right\}.
\end{align*}
Taking expectation with respect to $\mathcal G$ yields
\begin{align}
\label{supeq_Delta1_syn_term}
\mathbb P\left(
\left|
\frac{1}{\tilde n_1}
\sum_{i=1}^{\tilde n_1}
\ell_f(\tilde{\bm X}_i^{(1)},1)
-
\mathbb E_{\tilde{\mathcal P}_1}
\ell_f(\tilde{\bm X},1)
\right|>t
\right)
\leq
2\exp\left\{
-\frac{2\tilde n_1t^2}{M_\varepsilon^2}
\right\}.
\end{align}

Similarly, since the majority correction sample
$(\bm X_i)_{i\in\mathcal S_{0c}}$ consists of i.i.d. observations
from $\mathcal P_0$, Hoeffding's inequality gives
\begin{align}
\label{supeq_Delta0_1st_term}
\mathbb P\left(
\left|
\frac{1}{n_{0c}}
\sum_{i\in\mathcal S_{0c}}
\ell_f(\bm X_i,0)
-
\mathbb E_{\mathcal P_0}\ell_f(\bm X,0)
\right|>t
\right)
\leq
2\exp\left\{
-\frac{2n_{0c}t^2}{M_\varepsilon^2}
\right\}.
\end{align}

Finally, conditional on $\mathcal G$,
$\tilde{\bm X}_1^{(0)},\ldots,
\tilde{\bm X}_{\tilde n_0}^{(0)}$
are i.i.d. from $\tilde{\mathcal P}_0$, so
\begin{align*}
\mathbb P\left(
\left.
\left|
\frac{1}{\tilde n_0}
\sum_{i=1}^{\tilde n_0}
\ell_f(\tilde{\bm X}_i^{(0)},0)
-
\mathbb E_{\tilde{\mathcal P}_0}
\ell_f(\tilde{\bm X},0)
\right|>t
\,\right|\mathcal G
\right)
\leq
2\exp\left\{
-\frac{2\tilde n_0t^2}{M_\varepsilon^2}
\right\}.
\end{align*}
Integrating over $\mathcal G$ gives
\begin{align}
\label{supeq_Delta0_2nd_term}
\mathbb P\left(
\left|
\frac{1}{\tilde n_0}
\sum_{i=1}^{\tilde n_0}
\ell_f(\tilde{\bm X}_i^{(0)},0)
-
\mathbb E_{\tilde{\mathcal P}_0}
\ell_f(\tilde{\bm X},0)
\right|>t
\right)
\leq
2\exp\left\{
-\frac{2\tilde n_0t^2}{M_\varepsilon^2}
\right\}.
\end{align}

Let
\begin{align*}
t_1
&=
M_\varepsilon
\sqrt{\frac{\log(8/\alpha)}{2n_1^*}},
\qquad
t_2
=
M_\varepsilon
\sqrt{\frac{\log(8/\alpha)}{2\tilde n_1}},
\\
t_3
&=
M_\varepsilon
\sqrt{\frac{\log(8/\alpha)}{2n_{0c}}},
\qquad
t_4
=
M_\varepsilon
\sqrt{\frac{\log(8/\alpha)}{2\tilde n_0}}.
\end{align*}
Each of the four deviation events has probability at most
$\alpha/4$. Therefore, by a union bound, with probability at least
$1-\alpha$,
\begin{align}
\label{supeq_empirical_bounds}
|\widehat\Delta_1-\Delta_1|
\leq
t_1+t_2,
\qquad
|\widehat\Delta_0-\Delta_0|
\leq
t_3+t_4.
\end{align}

Combining
\eqref{supeq_Delta_decomposition},
\eqref{supeq_pop_Delta_bound}, and
\eqref{supeq_empirical_bounds}, we obtain, with probability at least
$1-\alpha$,
\begin{align*}
|\widehat\Delta_1-\widehat\Delta_0|
\leq
\varepsilon_{\rm BT}
+
M_\varepsilon
\sqrt{\frac{\log(8/\alpha)}{2}}
\left(
\frac{1}{\sqrt{n_1^*}}
+
\frac{1}{\sqrt{\tilde n_1}}
+
\frac{1}{\sqrt{n_{0c}}}
+
\frac{1}{\sqrt{\tilde n_0}}
\right).
\end{align*}
This proves the first claim.

For the second claim, if
$c_1\leq n_1/n\leq c_2<1/2$, then
$n_0-n_1=n-2n_1\asymp n$.
If, in addition,
\[
n_1^*,\tilde n_1,n_{0c},\tilde n_0
\asymp n_0-n_1,
\]
then there exists a constant $C<\infty$ such that
\begin{align*}
\frac{1}{\sqrt{n_1^*}}
+
\frac{1}{\sqrt{\tilde n_1}}
+
\frac{1}{\sqrt{n_{0c}}}
+
\frac{1}{\sqrt{\tilde n_0}}
\leq
\frac{C}{\sqrt{n_0-n_1}}.
\end{align*}
Thus, for any $\alpha\in(0,1)$, possibly after redefining $C$,
with probability at least $1-\alpha$,
\begin{align*}
|\widehat\Delta_1-\widehat\Delta_0|
\leq
\varepsilon_{\rm BT}
+
C\sqrt{
\frac{\log(8/\alpha)}
{n_0-n_1}
}.
\end{align*}
This completes the proof of
Proposition~\ref{prop_Delta0_minus_Delta1_upper_bound}.
\end{proof}

\subsection{Proof of Proposition~\ref{prop_Delta1_lower_bound}}
\begin{proof}
    We first show that there exists a constant $C > 0$ such that with probability tending to one,
    \begin{align}
        \label{supeq_conditional_l2}
        \mathbb E(U\|\bm X_J - \bm X_I\|_2\mid \mathcal D_1) \geq C\left(\frac{K}{n_1}\right)^{1/d}.
    \end{align}
    
    For each $i=1,\ldots,n_1$ and $k=1,\ldots,K$, let
    $\bm X_{i(k)}$ denote the $k$th nearest neighbour of $\bm X_i$
    in $\mathcal D_1\setminus\{\bm X_i\}$, and define
    $$
    R_{i,k}=\|\bm X_{i(k)}-\bm X_i\|_2.
    $$
    Since $\mathcal P_1$ is absolutely continuous, nearest-neighbour
    ties occur with probability zero.
    To establish the conditional lower bound in
    \eqref{supeq_conditional_l2}, we apply the law of large numbers for
    fixed-$K$ directed nearest-neighbour graphs established in
    Theorem~2.4 of \citet{penrose2003weak}.

    The graph-functional law of large numbers in Theorem~2.2 of \citet{penrose2003weak} requires the graph to be translation invariant, scale invariant, and stabilizing. For a fixed-$K$ directed nearest-neighbour graph, translating or positively rescaling all points preserves the nearest-neighbour relations, while rescaling by $b>0$ multiplies every edge length by $b$. Stabilization means that the edge contribution at a given point is determined by the observations within an almost surely finite random neighbourhood, thereby localizing the dependence among edge lengths. These properties are verified for the fixed-$K$ directed nearest-neighbour graph in the proof of their Theorem~2.4.
    
    The specialized version of their Theorem~2.4 can be stated as follows. Let $\mathcal D_m=\{\bm Z_1,\ldots,\bm Z_m\}$ consist of i.i.d. observations from a density $f$ on $\mathbb R^d$, and let
    $$
    L_K(\mathcal D_m)=\sum_{i=1}^m\sum_{k=1}^K\|\bm Z_{i(k)}-\bm Z_i\|_2
    $$
    be the total directed edge length of its fixed-$K$ nearest-neighbour graph. With $d>1$, if the following conditions hold for some $r>d/(d-1)$,
    $$
    \int_{\mathbb R^d}f(\bm x)^{1-1/d}\,{\rm d}\bm x<\infty
    \quad \text{ and }\quad
    \int_{\mathbb R^d}\|\bm x\|_2^r f(\bm x)\,{\rm d}\bm x<\infty,
    $$
    then
    $$
    m^{1/d-1}L_K(\mathcal D_m)\xrightarrow{L^1}\gamma_{d,K}\int_{\mathbb R^d}f(\bm x)^{1-1/d}\,{\rm d}\bm x,
    $$
    where $\gamma_{d,K}\in(0,\infty)$ depends only on $d$ and $K$.
    
    We apply this result with $m=n_1$ and $f=f_1$. Assumption~\ref{asmp_SMOTE_density}~(B2) ensures the remaining integrability conditions: since the support of $f_1$ is bounded, all polynomial moments are finite, including a moment of some order $r>d/(d-1)$. In addition, since the support has finite Lebesgue measure and $f_1$ is bounded, the integral of $f_1^{1-1/d}$ is finite and strictly positive.
    
    Consequently,
    $$
    n_1^{1/d-1}\sum_{i=1}^{n_1}\sum_{k=1}^K R_{i,k}\xrightarrow{\mathrm P}a_K,
    $$
    where
    $$
    a_K=\gamma_{d,K}\int_{{\rm supp}(\mathcal P_1)}f_1(\bm x)^{1-1/d}\,{\rm d}\bm x>0.
    $$
    Moreover, since $f_1(\bm x)\leq C_2$ on its support and
    $\int f_1(\bm x)\,{\rm d}\bm x=1$,
    \begin{align*}
    \int_{{\rm supp}(\mathcal P_1)}f_1(\bm x)^{1-1/d}\,{\rm d}\bm x
    &=\int_{{\rm supp}(\mathcal P_1)}
    f_1(\bm x)f_1(\bm x)^{-1/d}\,{\rm d}\bm x \\
    &\geq C_2^{-1/d}.
    \end{align*}
    Consequently,
    $$
    a_K\geq \gamma_{d,K}C_2^{-1/d}=: \underline a_K>0.
    $$
    Therefore, there exists a sequence $\eta_{K,n_1}\to0$ such that, with probability at least $1-\eta_{K,n_1}$,
    $$
    n_1^{1/d-1}\sum_{i=1}^{n_1}\sum_{k=1}^K R_{i,k}\geq\frac{\underline a_K}{2}.
    $$
    Equivalently,
    $$
    \frac{1}{n_1K}\sum_{i=1}^{n_1}\sum_{k=1}^K R_{i,k}\geq c_K\left(\frac{K}{n_1}\right)^{1/d},
    $$
    where
    $$
    c_K=\frac{\underline a_K}{2K^{1+1/d}}
    =\frac{\gamma_{d,K}C_2^{-1/d}}{2K^{1+1/d}}>0.
    $$
    Indeed, conditional on $\mathcal D_1$, the anchor index $I$ is uniform on $\{1,\ldots,n_1\}$, the neighbour rank $k$ is uniform on $\{1,\ldots,K\}$, $J=I(k)$, and $\mathbb E(U)=1/2$. Therefore, on the same event,
    \begin{align*}
    \mathbb E\{U\|\bm X_J-\bm X_I\|_2\mid\mathcal D_1\} &= \frac{1}{2n_1K}\sum_{i=1}^{n_1}\sum_{k=1}^K R_{i,k} \\
    &\geq \frac{c_K}{2}\left(\frac{K}{n_1}\right)^{1/d}.
    \end{align*}
    This establishes \eqref{supeq_conditional_l2} with $C=c_K/2$, where $\eta_{K,n_1}$ is the probability that the preceding nearest-neighbour lower-bound event fails.
    
    Denote $\mu_1 = \mathbb E_{\bm X\sim\mathcal P_1}\{\ell_f(\bm X,1)\}$ and
    $$
    \widehat\mu_1 =\frac{1}{n_1} \sum_{i=1}^{n_1} \ell_f(\bm X_i,1).
    $$
    Given the minority set $\mathcal D_1$, define the conditional expected loss under the SMOTE synthetic distribution as
    $$
    \tilde\mu_1(\mathcal D_1) = \mathbb E\{\ell_f(\tilde{\bm X},1)\mid\mathcal D_1\}.
    $$
    By Assumption~\ref{asmp_SMOTE_density}~(B3),
    \begin{align*}
        \left| \tilde\mu_1(\mathcal D_1)-\widehat\mu_1 \right|
        &=
        \left| \mathbb E[ \ell_f\{ \bm X_I+U(\bm X_J-\bm X_I),1\} - \ell_f(\bm X_I,1) \mid\mathcal D_1] \right|
        \\
        &\geq C_3 \mathbb E[ U\|\bm X_J-\bm X_I\|_2 \mid\mathcal D_1].
    \end{align*}
    Combining this inequality with the nearest-neighbor distance bound established above, with probability at least $1-\eta_{n_1}-\eta_{K,n_1}$,
    \begin{align*}
        \left|\tilde\mu_1(\mathcal D_1)-\widehat\mu_1\right|\geq c_L\left(\frac{K}{n_1}\right)^{1/d},
    \end{align*}
        where $c_L=C_3c_K/2>0$ depends only on $(d,C_2,C_3,K)$ and hence only on the constants listed in Proposition~\ref{prop_Delta1_lower_bound}.

    We next control the empirical fluctuations. Since $0\leq\ell_f(\bm X,1)\leq M_\varepsilon$ by Assumption~\ref{asmp:bounded}, Hoeffding's inequality gives, with probability at least $1-\alpha/3$,
    \begin{align*}
        \left|\widehat\mu_1-\mu_1\right|\leq M_\varepsilon\sqrt{\frac{\log(6/\alpha)}{2n_1}}.
    \end{align*}

    Let
    \begin{align*}
        \widehat\mu_1^*=\frac{1}{n_1^*}\sum_{i=1}^{n_1^*}\ell_f(\bm X_i^*,1),
    \end{align*}
    where $\bm X_1^*,\ldots,\bm X_{n_1^*}^*$ are independently generated from $\mathcal P_1$. Again by Hoeffding's inequality, with probability at least $1-\alpha/3$,
    \begin{align*}
        \left|\widehat\mu_1^*-\mu_1\right|\leq M_\varepsilon\sqrt{\frac{\log(6/\alpha)}{2n_1^*}}.
    \end{align*}

    Finally, define
    \begin{align*}
        \widehat{\tilde\mu}_1=\frac{1}{\tilde n_1}\sum_{j=1}^{\tilde n_1}\ell_f(\tilde{\bm X}_j,1).
    \end{align*}
    Conditional on $\mathcal D_1$, the synthetic samples $\tilde{\bm X}_1,\ldots,\tilde{\bm X}_{\tilde n_1}$ are independently generated from $\tilde{\mathcal P}_1(\mathcal D_1)$. Therefore, conditional on $\mathcal D_1$, Hoeffding's inequality gives that with probability at least $1-\alpha/3$,
    \begin{align*}
        \mathbb P\left[\left.\left|\widehat{\tilde\mu}_1-\tilde\mu_1(\mathcal D_1)\right|>M_\varepsilon\sqrt{\frac{\log(6/\alpha)}{2\tilde n_1}}\right|\mathcal D_1\right]\leq\frac{\alpha}{3}.
    \end{align*}
    Since the right hand side does not depend on the realization of $\mathcal D_1$, integrating over $\mathcal D_1$ yields
    \begin{align*}
        \mathbb P\left[\left|\widehat{\tilde\mu}_1-\tilde\mu_1(\mathcal D_1)\right|>M_\varepsilon\sqrt{\frac{\log(6/\alpha)}{2\tilde n_1}}\right]\leq\frac{\alpha}{3}.
    \end{align*}

    Recall that
    \begin{align*}
        \widehat\Delta_1=\widehat\mu_1^*-\widehat{\tilde\mu}_1.
    \end{align*}
    By adding and subtracting $\mu_1$, $\widehat\mu_1$, and $\tilde\mu_1(\mathcal D_1)$, and applying the reverse triangle inequality,
    \begin{align*}
        |\widehat\Delta_1|\geq\;\left|\widehat\mu_1-\tilde\mu_1(\mathcal D_1)\right|-\left|\widehat\mu_1-\mu_1\right|
        -\left|\widehat\mu_1^*-\mu_1\right|-\left|\widehat{\tilde\mu}_1-\tilde\mu_1(\mathcal D_1)\right|.
    \end{align*}
    Combining the preceding bounds and applying a union bound to the three concentration events, we obtain, with probability at least $1-\alpha-\eta_{n_1}-\eta_{K,n_1}$,
    \begin{align*}
        |\widehat\Delta_1|\geq\;c_L\left(\frac{K}{n_1}\right)^{1/d}-M_\varepsilon\sqrt{\frac{\log(6/\alpha)}{2n_1}}-M_\varepsilon\sqrt{\frac{\log(6/\alpha)}{2n_1^*}}-M_\varepsilon\sqrt{\frac{\log(6/\alpha)}{2\tilde n_1}}.
    \end{align*}
    The union bound does not require independence among these deviation events. This completes the proof of Proposition~\ref{prop_Delta1_lower_bound}.

\end{proof}

\subsection{Proof of Theorem~\ref{thm_bc_ERM}}
\begin{proof}
Throughout this proof, we condition on the realized class labels, so that $n_1$ and $n_0$ are fixed. Note that
\begin{align*}
    L^*(\widehat f^{\rm bc}) - L^*(f^*) = & \left(L^*(\widehat f^{\rm bc}) - L^{\rm bc}(\widehat f^{\rm bc})\right) + \left(L^{\rm bc}(\widehat f^{\rm bc}) - L^{\rm bc}(f^*)\right) + \left(L^{\rm bc}(f^*) - L^*(f^*)\right),
\end{align*}
and \begin{align*}
    L^{\rm bc}(\widehat f^{\rm bc}) - L^{\rm bc}(f^*)\leq 0.
\end{align*}
We just need to show that for any prediction function $f\in\mathcal F$ and any $\alpha \in (0, 1)$, with conditional probability at least $1-\alpha$,
\begin{align}
    \label{supeq_Lbc_Lbal_error}
    \begin{split}
    & \sup_{f\in\mathcal F}|L^{\rm bc}(f) - L^*(f)|\\
    \leq & \frac{n_0-n_1}{2n_0}\varepsilon_{\rm BT} + \frac{C}{2}\cdot(2C_{\mathcal F}+\sqrt{\log(12/\alpha)})\left\{\frac{1}{\sqrt{n_1}} + \frac{1}{\sqrt{n_0}} + \frac{1}{\sqrt{n_{0c}}} + \frac{1}{\sqrt{\tilde n_1}} + \frac{1}{\sqrt{\tilde n_0}}\right\}.
    \end{split}
\end{align}
Recall that 
\begin{align*}
    L^{\rm bc}(f) =& \frac{1}{n+\tilde n_1}\left[\sum_{i=1}^n \ell_f(\bm X_i, Y_i) + \tilde n_1\left\{\frac{1}{\tilde n_1}\sum_{i=1}^{\tilde n_1}\ell_f(\tilde{\bm X}_i^{(1)}, 1) + \widehat\Delta_0\right\}\right],
    \\
    L^*(f) = & \frac{1}{2}\mathbb E_{\mathcal P_1}[\ell_f(\bm X, 1)] + \frac{1}{2}\mathbb E_{\mathcal P_0}[\ell_f(\bm X, 0)].
\end{align*}
First, rewrite $L^{\rm bc}(f) - L^*(f)$ as 
\begin{align*}
    L^{\rm bc}(f) - L^*(f) =& \underbrace{\frac{n_1}{n+\tilde n_1}\cdot \frac{1}{n_1}\sum_{i=1}^{n_1}\ell_f(\bm X_i, 1) - \left(\frac{1}{2} - \frac{\tilde n_1}{n+\tilde n_1}\right)\cdot\mathbb E_{\mathcal P_1}[\ell_f(\bm X, 1)]}_{(\rm I)}
    \\
    &+ \underbrace{\frac{n_0}{n+\tilde n_1}\cdot \frac{1}{n_0}\sum_{i=n_1+1}^{n}\ell_f(\bm X_i, 0) - \frac{1}{2}\cdot\mathbb E_{\mathcal P_0}[\ell_f(\bm X, 0)]}_{(\rm II)}
    \\
    &+ \frac{\tilde n_1}{n+\tilde n_1}\left\{\frac{1}{\tilde n_1}\sum_{i=1}^{\tilde n_1}\ell_f(\tilde{\bm X}_i^{(1)}, 1) + \Delta_0 + \underbrace{(\widehat\Delta_0 - \Delta_0)}_{(\rm III)} - \mathbb E_{\mathcal P_1}[\ell_f(\bm X, 1)]\right\}.
\end{align*}

By the definitions of $\Delta_1(f)$ and $\Delta_0(f)$,
\begin{align*}
    &\frac{1}{\tilde n_1}\sum_{i=1}^{\tilde n_1}\ell_f(\tilde{\bm X}_i^{(1)},1)
    +\Delta_0(f)-\mathbb E_{\mathcal P_1}\ell_f(\bm X,1)
    \\
    ={}&\underbrace{\frac{1}{\tilde n_1}\sum_{i=1}^{\tilde n_1}\ell_f(\tilde{\bm X}_i^{(1)},1)
    -\mathbb E_{\tilde{\mathcal P}_1}\ell_f(\tilde{\bm X},1)}_{(\rm IV)}
    +\underbrace{\Delta_0(f)-\Delta_1(f)}_{(\rm V)}.
\end{align*}
Consequently,
\begin{align*}
    L^{\rm bc}(f)-L^*(f)
    =(\rm I)+(\rm II)
    +\frac{\tilde n_1}{n+\tilde n_1}\left\{(\rm III)+(\rm IV)+(\rm V)\right\}.
\end{align*}
We now bound these terms uniformly over $f\in\mathcal F$.

\begin{lemma}[Uniform concentration for bounded losses]
    \label{lemma:uniform_concentration}
    Under Assumption~\ref{asmp:bounded}, let $Q$ be a probability distribution on $\mathcal X\times\{0,1\}$ satisfying $\mathfrak R_m(\mathcal L;Q)\leq C_{\mathcal F}/\sqrt m$. For any i.i.d. sample $\bm Z_1,\dots,\bm Z_m$ from $Q$ and any $\delta\in(0,1)$, with probability at least $1-\delta$,
    \begin{align}
        \label{eq:uniform_concentration}
        \sup_{f\in\mathcal F}\left|\frac{1}{m}\sum_{i=1}^m\ell_f(\bm Z_i)-\mathbb E_Q\ell_f(\bm Z)\right|\le M_\varepsilon\left(\frac{2C_{\mathcal F}}{\sqrt m}+\sqrt{\frac{\log(2/\delta)}{2m}}\right).
    \end{align}
\end{lemma}
\begin{proof}[Proof of Lemma \ref{lemma:uniform_concentration}]
    By Assumption \ref{asmp:bounded}, $0\le\ell_f(\bm z)\le M_\varepsilon$ for all $\bm z$ and all $f\in\mathcal F$. Define $g_f(\bm z)=\ell_f(\bm z)/M_\varepsilon$ and $\mathcal L=\{g_f:f\in\mathcal F\}$, then $g_f(\bm z)\in[0,1]$.

    Define
    \begin{align*}
        \Phi(\bm Z_{1:m})=\sup_{g\in\mathcal L}\left|\frac{1}{m}\sum_{i=1}^m g(\bm Z_i)-\mathbb E_Qg(\bm Z)\right|.
    \end{align*}
    We first control $\mathbb E\Phi(\bm Z_{1:m})$ by symmetrization and Rademacher complexity. Let $\bm Z_1',\dots,\bm Z_m'$ be i.i.d. copies independent of $\bm Z_{1:m}$. Then
    \begin{align*}
        \mathbb E\Phi(\bm Z_{1:m})
        &=\mathbb E\sup_{g\in\mathcal L}\left|\frac{1}{m}\sum_{i=1}^m g(\bm Z_i)-\mathbb E_Qg(\bm Z)\right| \\
        &=\mathbb E\sup_{g\in\mathcal L}\left|\frac{1}{m}\sum_{i=1}^m \left(g(\bm Z_i)-\mathbb E_Qg(\bm Z_i')\right)\right| \\
        &\le \mathbb E\sup_{g\in\mathcal L}\left|\frac{1}{m}\sum_{i=1}^m\left(g(\bm Z_i)-g(\bm Z_i')\right)\right|.
    \end{align*}
    Let $\sigma_1,\dots,\sigma_m$ be i.i.d. Rademacher random variables independent of all random variables. Conditioning on $\bm Z_{1:m}$ and $\bm Z_{1:m}'$, the standard symmetrization step gives
    \begin{align*}
        \mathbb E\sup_{g\in\mathcal L}\left|\frac{1}{m}\sum_{i=1}^m\left(g(\bm Z_i)-g(\bm Z_i')\right)\right|
        =& \mathbb E\sup_{g\in\mathcal L}\left|\frac{1}{m}\sum_{i=1}^m \sigma_i\left(g(\bm Z_i)-g(\bm Z_i')\right)\right|
        \\
        \le& 2\mathbb E\sup_{g\in\mathcal L}\left|\frac{1}{m}\sum_{i=1}^m\sigma_i g(\bm Z_i)\right|
        =2\mathfrak R_m(\mathcal L;Q).
    \end{align*}
    Therefore $\mathbb E\Phi(\bm Z_{1:m})\le 2\mathfrak R_m(\mathcal L;Q)$.

    Next we apply McDiarmid's inequality to $\Phi(\bm Z_{1:m})$. Fix $j\in\{1,\dots,m\}$ and replace $\bm Z_j$ by an independent copy $\bm Z_j'$. Denote by $\bm Z_{1:m}^{(j)}$ the sample with only the $j$th coordinate replaced. Since $g(\bm z)\in[0,1]$, for any $g\in\mathcal L$,
    \begin{align*}
        \left|\frac{1}{m}\sum_{i=1}^m g(\bm Z_i)-\frac{1}{m}\sum_{i=1}^m g\left(\bm Z_i^{(j)}\right)\right|\le\frac{1}{m}.
    \end{align*}
    Taking supremum over $g\in\mathcal L$ and using the inequality $|\sup a-\sup b|\le\sup|a-b|$ yields
    \begin{align*}
        \left|\Phi(\bm Z_{1:m})-\Phi\left(\bm Z_{1:m}^{(j)}\right)\right|\le\frac{1}{m}.
    \end{align*}
    Thus $\Phi$ satisfies the bounded differences condition with constants $c_j=1/m$ for all $j$. By McDiarmid's inequality, for any $\delta\in(0,1)$, with probability at least $1-\delta$,
    \begin{align*}
        \Phi(\bm Z_{1:m})\le\mathbb E\Phi(\bm Z_{1:m})+\sqrt{\frac{\log(2/\delta)}{2}\sum_{j=1}^m c_j^2}
        =\mathbb E\Phi(\bm Z_{1:m})+\sqrt{\frac{\log(2/\delta)}{2m}}.
    \end{align*}
    Combining this with $\mathbb E\Phi(\bm Z_{1:m})\le 2\mathfrak R_m(\mathcal L;Q)$ and the assumed bound on $\mathfrak R_m(\mathcal L;Q)$ gives
    \begin{align*}
        \Phi(\bm Z_{1:m})\le \frac{2C_{\mathcal F}}{\sqrt m}+\sqrt{\frac{\log(2/\delta)}{2m}}
    \end{align*}
    with probability at least $1-\delta$. Finally, multiplying both sides by $M_\varepsilon$ and using $\ell_f=M_\varepsilon g_f$ yields \eqref{eq:uniform_concentration}.
\end{proof}

For $m\geq1$ and $\delta\in(0,1)$, define
$$
b_m(\delta)
=M_\varepsilon\left\{
\frac{2C_{\mathcal F}}{\sqrt m}
+\sqrt{\frac{\log(2/\delta)}{2m}}
\right\}.
$$
For the synthetic observations, Lemma~\ref{lemma:uniform_concentration}
is applied conditionally on $\mathcal G$. Here and below,
$\mathbb E_{\tilde{\mathcal P}_y}$ denotes the conditional expectation
$\mathbb E_{\tilde{\mathcal P}_y(\mathcal G)}$ and is therefore
$\mathcal G$-measurable. By Assumptions~\ref{asmp:complexity} and
\ref{asmp:synthetic}, for each $y\in\{0,1\}$,
\begin{align}
&\mathbb P\Bigg[
\left.
\sup_{f\in\mathcal F}
\left|
\frac{1}{\tilde n_y}
\sum_{i=1}^{\tilde n_y}
\ell_f(\tilde{\bm X}_i^{(y)},y)
-\mathbb E_{\tilde{\mathcal P}_y(\mathcal G)}
\ell_f(\tilde{\bm X},y)
\right|
>b_{\tilde n_y}(\delta)
\,\right|\,\mathcal G
\Bigg]
\leq\delta
\quad\text{almost surely}.
\label{eq:conditional_uniform_concentration}
\end{align}
Writing $\mathcal E_y(\delta)$ for the exceedance event inside
\eqref{eq:conditional_uniform_concentration}, the tower property gives
\begin{align*}
\mathbb P\{\mathcal E_y(\delta)\}
&=\mathbb E\left[\mathbb P\{\mathcal E_y(\delta)\mid\mathcal G\}\right]
\leq\delta.
\end{align*}
Thus the same deviation bound also holds unconditionally.

Applying \eqref{eq:uniform_concentration} with
$Q=\mathcal P_1\otimes\delta_1$ and
$Q=\mathcal P_0\otimes\delta_0$ to the minority and majority empirical averages, respectively, and taking $\delta=\alpha/6$ in each application, we have with probability at least $1-\alpha/6$ for each,
\begin{align*}
    \sup_{f\in\mathcal F}\left|\frac{1}{n_1}\sum_{i=1}^{n_1}\ell_f(\bm X_i,1)-\mathbb E_{\mathcal P_1}\ell_f(\bm X,1)\right|&\le M_\varepsilon\left(\frac{2C_{\mathcal F}}{\sqrt{n_1}}+\sqrt{\frac{\log(12/\alpha)}{2n_1}}\right),
    \\
    \sup_{f\in\mathcal F}\left|\frac{1}{n_0}\sum_{i=n_1+1}^{n}\ell_f(\bm X_i,0)-\mathbb E_{\mathcal P_0}\ell_f(\bm X,0)\right|&\le M_\varepsilon\left(\frac{2C_{\mathcal F}}{\sqrt{n_0}}+\sqrt{\frac{\log(12/\alpha)}{2n_0}}\right).
\end{align*}
Therefore,
\begin{align*}
    \sup_{f\in\mathcal F}|({\rm I})|&\le \frac{n_1}{n+\tilde n_1}M_\varepsilon\left(\frac{2C_{\mathcal F}}{\sqrt{n_1}}+\sqrt{\frac{\log(12/\alpha)}{2n_1}}\right)+\left|\frac{n_1 + \tilde n_1}{n+\tilde n_1}-\frac{1}{2}\right|M_\varepsilon,\\
    \sup_{f\in\mathcal F}|({\rm II})|&\le \frac{n_0}{n+\tilde n_1}M_\varepsilon\left(\frac{2C_{\mathcal F}}{\sqrt{n_0}}+\sqrt{\frac{\log(12/\alpha)}{2n_0}}\right)+\left|\frac{n_0}{n+\tilde n_1}-\frac{1}{2}\right|M_\varepsilon.
\end{align*}

Recall that
\begin{align*}
    ({\rm III}) =\left(\frac{1}{n_{0c}}\sum_{i\in\mathcal S_{0c}}\ell_f(\bm X_i,0)-\mathbb E_{\mathcal P_0}\ell_f(\bm X,0)\right)-\left(\frac{1}{\tilde n_0}\sum_{i=1}^{\tilde n_0}\ell_f(\tilde{\bm X}_i^{(0)},0)-\mathbb E_{\tilde{\mathcal P}_0}\ell_f(\tilde{\bm X},0)\right).
\end{align*}
Applying \eqref{eq:uniform_concentration} to the real correction sample and \eqref{eq:conditional_uniform_concentration} with $y=0$ to the synthetic majority sample, taking $\delta=\alpha/6$ in each application, and then applying a union bound, we have with probability at least $1-\alpha/3$,
\begin{align*}
    \sup_{f\in\mathcal F}|({\rm III})|\le M_\varepsilon\left(\frac{2C_{\mathcal F}}{\sqrt{n_{0c}}}+\sqrt{\frac{\log(12/\alpha)}{2n_{0c}}}\right)+M_\varepsilon\left(\frac{2C_{\mathcal F}}{\sqrt{\tilde n_0}}+\sqrt{\frac{\log(12/\alpha)}{2\tilde n_0}}\right).
\end{align*}

Applying \eqref{eq:conditional_uniform_concentration} with $y=1$ and $\delta=\alpha/6$, followed by the tower-property argument above, we have with probability at least $1-\alpha/6$,
\begin{align*}
    \sup_{f\in\mathcal F}|({\rm IV})|\le M_\varepsilon\left(\frac{2C_{\mathcal F}}{\sqrt{\tilde n_1}}+\sqrt{\frac{\log(12/\alpha)}{2\tilde n_1}}\right).
\end{align*}

By Assumption~\ref{asmp:bias_transfer},
\begin{align*}
    \sup_{f\in\mathcal F}|({\rm V})|
    =\sup_{f\in\mathcal F}|\Delta_0(f)-\Delta_1(f)|
    \leq\varepsilon_{\rm BT}.
\end{align*}

By the triangle inequality, 
\begin{align*}
    \sup_{f\in\mathcal F}|L^{\rm bc}(f)-L^*(f)|\le&\sup_{f\in\mathcal F}|({\rm I})|+\sup_{f\in\mathcal F}|({\rm II})|\\
    &+\frac{\tilde n_1}{n+\tilde n_1}\left(\sup_{f\in\mathcal F}|({\rm III})| + \sup_{f\in\mathcal F}|({\rm IV})|+\sup_{f\in\mathcal F}|({\rm V})|\right).
\end{align*}

Under the assumption that $\tilde n_1=n_0-n_1$, we have $n+\tilde n_1=2n_0$ and
\begin{align*}
    \frac{n_1}{n+\tilde n_1} = \frac{n_1}{2n_0}, \quad
    \frac{1}{2} - \frac{\tilde n_1}{n+\tilde n_1} = \frac{n_1}{2n_0}, \quad
    \frac{n_0}{n+\tilde n_1} = \frac{1}{2}, \quad \text{and}\quad
    \frac{\tilde n_1}{n+\tilde n_1}=\frac{n_0-n_1}{2n_0}.
\end{align*} 
Collecting the uniform bounds for $({\rm I})$--$({\rm V})$ and applying a union bound over the five concentration events above, with probability at least $1-5\alpha/6$, we obtain
\begin{align*}
    \sup_{f\in\mathcal F}|L^{\rm bc}(f)-L^*(f)|
    \le& M_\varepsilon\left\{\frac{n_1}{2n_0}\left(\frac{2C_{\mathcal F}}{\sqrt{n_1}}+\sqrt{\frac{\log(12/\alpha)}{2n_1}}\right)+\frac{1}{2}\left(\frac{2C_{\mathcal F}}{\sqrt{n_0}}+\sqrt{\frac{\log(12/\alpha)}{2n_0}}\right)\right.
    \\
    &\left.+\frac{n_0-n_1}{2n_0}\left(\frac{2C_{\mathcal F}}{\sqrt{n_{0c}}}+\sqrt{\frac{\log(12/\alpha)}{2n_{0c}}}+\frac{2C_{\mathcal F}}{\sqrt{\tilde n_0}}+\sqrt{\frac{\log(12/\alpha)}{2\tilde n_0}}\right)\right.
    \\
    &\left.+\frac{n_0-n_1}{2n_0}\left(\frac{2C_{\mathcal F}}{\sqrt{\tilde n_1}}+\sqrt{\frac{\log(12/\alpha)}{2\tilde n_1}}\right)\right\} + \frac{n_0-n_1}{2n_0}\varepsilon_{\rm BT}.
\end{align*}
The five concentration events need not be independent; the stated probability follows directly from the union bound.
Since $c_1\le n_1/n\le c_2$, the coefficients in the preceding display are uniformly bounded. Hence, after redefining $C$, conditional on the realized class labels, with probability at least $1-5\alpha/6\ge 1-\alpha$,
\begin{align*}
    \sup_{f\in\mathcal F}|L^{\rm bc}(f)-L^*(f)|
    \leq& \frac{n_0-n_1}{2n_0}\varepsilon_{\rm BT}
    + \frac{C}{2}(2C_{\mathcal F}+\sqrt{\log(12/\alpha)})
    \left\{\frac{1}{\sqrt{n_1}}+\frac{1}{\sqrt{n_0}}+\frac{1}{\sqrt{n_{0c}}}\right.\\
    &\left.\hspace{46mm}+\frac{1}{\sqrt{\tilde n_1}}+\frac{1}{\sqrt{\tilde n_0}}\right\},
\end{align*}
where $C<\infty$ depends only on $M_\varepsilon,c_1,c_2$. This proves \eqref{supeq_Lbc_Lbal_error}.

Finally, we have
\begin{align*}
    L^*(\widehat f^{\rm bc}) - L^*(f^*) = & \left(L^*(\widehat f^{\rm bc}) - L^{\rm bc}(\widehat f^{\rm bc})\right) + \left(L^{\rm bc}(\widehat f^{\rm bc}) - L^{\rm bc}(f^*)\right) + \left(L^{\rm bc}(f^*) - L^*(f^*)\right)
    \\
    \leq & \frac{n_0-n_1}{n_0}\varepsilon_{\rm BT}
    + C(2C_{\mathcal F}+\sqrt{\log(12/\alpha)})
    \left\{\frac{1}{\sqrt{n_1}}+\frac{1}{\sqrt{n_0}}+\frac{1}{\sqrt{n_{0c}}}\right.\\
    &\left.\hspace{43mm}+\frac{1}{\sqrt{\tilde n_1}}+\frac{1}{\sqrt{\tilde n_0}}\right\}
\end{align*}
with conditional probability at least $1-\alpha$, where the inequality holds since $L^{\rm bc}(\widehat f^{\rm bc}) - L^{\rm bc}(f^*) \leq 0$. This completes the proof of Theorem~\ref{thm_bc_ERM}.
\end{proof}

\section{Numerical Studies}\label{sec_simu}
In this section, we present simulation results under a range of experimental settings. Theorem~\ref{thm_bc_ERM} considers the exactly balanced augmentation regime $\tilde n_1=n_0-n_1$. Some experiments below deliberately use lighter augmentation and should therefore be viewed as complementary empirical evaluations rather than direct numerical verifications of that theorem.

\subsection{Mean Shift Model}
The simulation investigates the performance of imbalanced classification on synthetic augmented data with and without bias correction. We consider an extremely imbalanced binary classification problem with sample size $n=10{,}000$ and feature dimension $d=10$.
For each replicate, labels $Y\in\{0,1\}$ are generated i.i.d.\ from $\mathrm{Bernoulli}(\pi_1)$ with minority prevalence $\pi_1=0.005$.
Conditional on $Y$, covariates are generated with independent coordinates and a class-dependent location shift: $\bm X\mid(Y=0)$ is centered at the origin, whereas $\bm X\mid(Y=1)$ is shifted by $\bm\mu$, where $\bm\mu=\mathbf{1}_d$ for the $t(2)$ and logistic settings, and $\mu=0.5\,\mathbf{1}_d$ for the normal setting.
We examine three distributional settings for the feature noise: (i) heavy-tailed Student-$t$ with $2$ degrees of freedom, i.e., $\bm X=\bm \xi+\bm\mu\,I(Y=1)$ with elements $\xi_j\stackrel{\text{i.i.d.}}{\sim}t(2)$; (ii) Gaussian, with $\xi_j\stackrel{\text{i.i.d.}}{\sim}\mathcal{N}(0,1)$; and (iii) logistic, with $\xi_j\stackrel{\text{i.i.d.}}{\sim}\mathrm{Logistic}(0,1)$.
Each replicate is split into training, validation and test sets with proportions $60\%/20\%/20\%$.
We fit logistic regression models using gradient descent for $100$ epochs with learning rate $0.05$.
For methods using augmentation, we generate $n_s=\lfloor (n_0-n_1)\cdot \texttt{syn\_rate}\rfloor$ additional minority samples, where $n_0$ and $n_1$ are the majority/minority counts in the training set and $\texttt{syn\_rate}=2\pi_1(1-2\pi_1)$. With $\pi_1=0.005$, $\texttt{syn\_rate}\approx 0.0099$, so the augmented training data remain highly imbalanced. This setting deliberately departs from the balanced regime $\tilde n_1=n_0-n_1$ considered in the theoretical analysis and instead examines the behaviour of the correction under light augmentation. We use $k=15$ nearest neighbors for SMOTE.
As a non-synthetic baseline for imbalance, we also consider sample-size reweighting via a weighted logistic loss that upweights minority observations.

Consider the loss function $\ell(\bm x, y;\bm\beta) = -y\cdot \bm x^\top\bm\beta + \log(1+\exp(\bm x^\top\bm\beta))$. For the three methods, using raw data, synthetic-augmented data, and synthetic-augmented data with bias correction, we train models by minimizing the respective loss functions given in Equations~(\ref{eq_raw_loss}), (\ref{eq_syn_loss}), and (\ref{eq_bc_loss}) for 100 epochs. 

Due to the imbalanced nature of the simulated data, overall accuracy is not an appropriate performance metric and can be misleading, particularly when class proportions are modified through upsampling.
Therefore, we evaluate all methods using metrics that are more informative for minority-class performance, including recall, precision, and $F_\beta$-score with $\beta=1/\pi_1$. The $F_\beta$ score is defined as 
\begin{align*}
    F_\beta = \frac{(1+\beta^2){\rm TP}}{(1+\beta^2){\rm TP} + {\rm FP} + \beta^2 {\rm FN}},
\end{align*}
where ${\rm TP}, {\rm FP}$ and ${\rm FN}$ represent the number of true positives, false positives and false negatives, respectively. In this case we choose $\beta = 1/\pi_1$, to account for class imbalance, so that false negatives are penalized more heavily when the minority class is rare. Importantly, all upsampling procedures are applied exclusively to the training samples. Performance is evaluated on test sets that preserve the original class distribution, so that the reported results reflect performance under the true data-generating mechanism rather than the resampled training space.

\begin{table}[htbp]
    \centering
    \caption{Performance metrics (recall, precision, $F_\beta$-score with $\beta=1/\pi_1$) evaluated on raw and synthetic-augmented data, with/without bias correction (denoted as SMOTE and BC, respectively), across varying distributions based on the mean shift model. The results are based on 100 simulations and bold values indicate the top-performing method per metric.}
    \resizebox{\textwidth}{!}{ 
        \begin{tabular}{cccccccccc}
\multicolumn{1}{l}{} & \multicolumn{3}{c}{$t(2)$}                          & \multicolumn{3}{c}{$\mathcal{N}(\cdot,1)$}          & \multicolumn{3}{c}{$Logistic(\cdot,1)$}             \\
\multicolumn{1}{l}{} & Recall          & Precision       & $F_\beta$ Score & Recall          & Precision       & $F_\beta$ Score & Recall          & Precision       & $F_\beta$ Score \\ \hline
Raw                  & 0.5942          & 0.0059          & 0.5927          & 0.6404          & 0.0065          & 0.6388          & 0.6384          & 0.0065          & 0.6368          \\
SMOTE                & 0.7240          & 0.0071          & 0.7221          & 0.7916          & \textbf{0.0080} & 0.7896          & 0.7927          & \textbf{0.0081} & 0.7907          \\
Bootstrap            & 0.7163          & 0.0071          & 0.7145          & 0.7825          & 0.0079          & 0.7806          & 0.7931          & 0.0081          & 0.7912          \\
Reweight             & 0.6703          & 0.0066          & 0.6686          & 0.7271          & 0.0073          & 0.7253          & 0.7196          & 0.0073          & 0.7178          \\
SMOTE-BC             & \textbf{0.7317} & \textbf{0.0072} & \textbf{0.7298} & \textbf{0.7962} & \textbf{0.0080} & \textbf{0.7942} & \textbf{0.7972} & \textbf{0.0081} & \textbf{0.7952}
\end{tabular}
    }
    \label{lab:recallprecisionF1}
\end{table}

The resulting evaluation metrics for the three methods are summarized in Table~\ref{lab:recallprecisionF1}. Across the three distributions and 100 simulations, SMOTE improved recall, precision, and the $F_\beta$-score relative to the raw-data analysis. Adding bias correction yielded further improvements in these metrics in the settings considered.

\subsection{Non-linear Classification}

\begin{figure}[htbp]
  \centering
  \includegraphics[width=\linewidth]{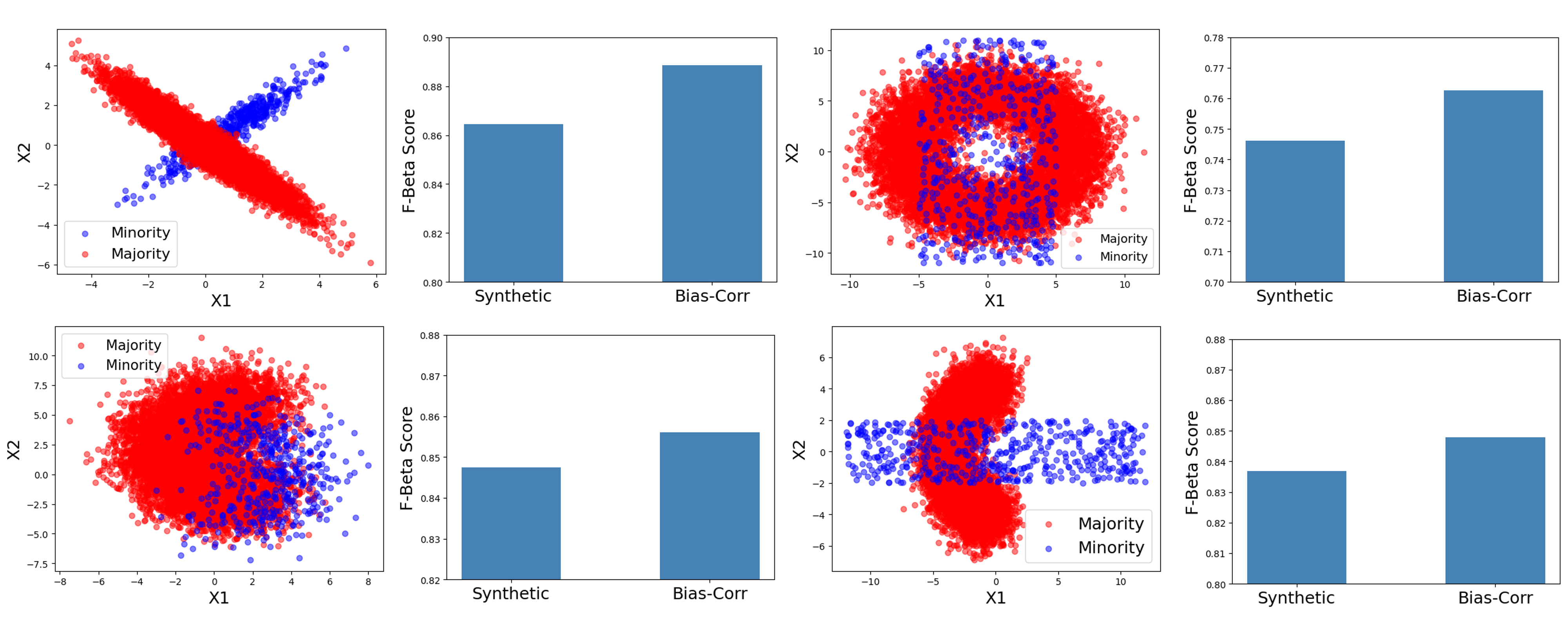}
  \caption{Distribution and $F_\beta$ score for four non-linear classification settings.}
  \label{nonlinear}
\end{figure}

To evaluate the proposed bias correction technique under diverse non-linear decision boundaries, we consider four controlled two-dimensional classification settings, each representing a distinct geometric relationship between the majority and minority classes as illustrated in Figure~\ref{nonlinear}. In each replicate, we generate $n=1,000$ observations with $d=2$ features and labels $Y\in\{0,1\}$ sampled i.i.d.\ from $\mathrm{Bernoulli}(\pi_1)$ with minority prevalence $\pi_1=0.05$.
The four configurations vary in class overlap, non-convexity, and variance structure: one setting produces two elongated clouds with opposing correlation directions, yielding intersecting diagonal structures, another places the majority class on a noisy ring surrounding a minority strip, a third yields overlapping clusters with unequal spread and a location shift, and a fourth forms a clearly non-convex majority distribution with two lobes while the minority class occupies a horizontal band.
For evaluation, we generate an independent test set of size $0.5n$ with balanced labels $\mathrm{Bernoulli}(0.5)$ to assess performance without class-prior effects.
For each configuration, we generate synthetic samples using SMOTE with $k=5$ nearest neighbors and $n_s=\lfloor (n_0-n_1)\cdot \texttt{syn\_rate}\rfloor$ (with $\texttt{syn\_rate}=0.5$), and then compare standard synthetic augmentation against the bias-corrected objective.
We train the classifier using a two-hidden-layer ReLU network with 12 units per layer, optimized by SGD for 500 epochs (learning rate $0.05$ and weight decay $10^{-4}$).
Performance is summarized by the $F_\beta$ score with $\beta=1/\pi_1$, averaged over 100 independent runs.

For each configuration, we compare standard synthetic augmentation with the bias-corrected objective using the same classifier. Performance is evaluated on an independent balanced test set using the $F_\beta$ score, with results averaged over $100$ replicates. 

Figure~\ref{nonlinear} illustrates the data distributions and corresponding $F_\beta$ scores for four non-linear classification settings. Across these configurations, adding bias correction improved the performance of synthetic data augmentation. The improvement was also observed in the two settings with non-convex majority support, where convex interpolation by SMOTE need not preserve the geometry of the underlying distribution, and in settings with high minority variance. These results provide empirical evidence for the correction in the configurations considered.

\subsection{Sigmoid Bernoulli Model}
In this section, a simulation study on the sigmoid Bernoulli model is conducted to evaluate the performance of the proposed bias correction technique across several data-generating distributions and synthetic generators.
Each replicate uses sample size $n=1{,}000$ and dimension $d=10$.
Covariates $\bm X\in\mathbb{R}^d$ are generated from one of several parametric families with a location vector $\mu$ and a common scale factor, and responses follow a sigmoid Bernoulli model
\[
\mathbb{P}(Y=1\mid \bm X)=\sigma(\bm X^\top \bm \beta^\star), \qquad \bm \beta^\star=-\mathbf{1}_d,
\]
where $\sigma(\cdot)$ denotes the logistic sigmoid function.
We mainly study four location--scale families for elements of $\bm X$: Gaussian, Gumbel, location-scale $t$ (with $t_4$ noise), and the hyperbolic secant distribution (HSD), and we additionally test the Laplace and logistic distributions to assess robustness.
For each distribution, $\bm \mu$ is chosen to induce a highly imbalanced label distribution (with a minority rate around $\pi_1=0.05$), and the data are split into training/validation/test sets with proportions $60\%/20\%/20\%$.
To address imbalance, use SMOTE with $k=5$ nearest neighbors and compare it with the proposed bias correction method.
The number of synthetic samples is set to $\tilde n=\lfloor (n_0-n_1)\cdot \texttt{syn\_rate}\rfloor$ with $\texttt{syn\_rate}=10\pi_1(1-2\pi_1)$.
Performance is evaluated by the estimation error of $\bm \beta$, averaged over $100$ independent simulation runs.

\begin{table}[htbp]
    \centering
    \caption{Estimation error of parameter $\bm \beta$ evaluated on raw and synthetic data (with/without bias correction) across varying distributions based on sigmoid Bernoulli model. The results are based on 100 simulations and bold values indicate the top-performing method.}
        \begin{tabular}{ccccccc}
                     & Gaussian                    & Gumbel                        & Loc-Scale t                            & HSD                          & Laplace                  & Logistic                 \\ \hline
Raw                  & 2.441                       & 2.426                         & 2.431                                  & 2.496                        & 2.401                    & 2.384                    \\
SMOTE                & 2.310                       & 2.311                         & 2.347                                  & 2.380                        & 2.330                    & 2.362                    \\
Bias Corr            & \textbf{2.299}              & \textbf{2.308}                & \textbf{2.332}                         & \textbf{2.369}               & \textbf{2.318}           & \textbf{2.350}
     \\\hline
\end{tabular}
    \label{lab:betaMSE}
\end{table}

Table~\ref{lab:betaMSE} reports the parameter estimation error for $\beta$ under the sigmoid Bernoulli model. SMOTE reduced the error relative to the raw-data analysis across the tested distributions, and adding bias correction produced a further reduction. Thus, in these simulation settings, bias correction improved recovery of the underlying coefficient vector beyond standard synthetic augmentation.

\subsection{Sensitivity to the Bias-Transfer Assumption}

We conduct a sensitivity simulation to examine how the performance of bias correction changes as the bias-transfer relationship between the majority and minority classes gradually deteriorates. The simulation is designed so that the two class-specific synthetic biases are closely matched under the baseline setting, while the minority distribution is progressively perturbed to increase the bias-transfer discrepancy.

We consider $d=10$ covariates with $n_0=9500$ majority observations and $n_1=500$ minority observations. Consider the majority and minority distributions as
$$
\mathcal P_0
=
\frac{1}{2}N(-\mu-4e_1,\Sigma)
+
\frac{1}{2}N(-\mu+4e_1,\Sigma),
$$
and
$$
\mathcal P_{1,\gamma}
=
\frac{1}{2}N(\mu-4e_1,\Sigma)
+
\frac{1}{2}N\{\mu+(4+12\gamma)e_1,\Sigma\},
$$
where
$$
\Sigma_{jk}
=
0.72(0.42)^{|j-k|},\quad 
\mu
=
(2.60,0.45,0.35,0.30,0.25,0.20,0.15,0.10,0.08,0.05)^\top,
$$
and
$$
\gamma\in\{0,0.05,0.10,0.15,0.20,0.25,0.5,0.75,1\}.
$$
Thus the majority distribution is fixed and the minority distribution is changed via $\gamma$ to progressively violate this reflection relationship.

For each class, synthetic observations are generated from a separately fitted diagonal Gaussian distribution $N\{\widehat\mu_y,\operatorname{diag}(\widehat\Sigma_y)\}.$
The majority sample is equally divided into a generation set and a correction set, with $n_{0g}=n_{0c}=4750$. We generate $\tilde n_1=n_0-n_1=9000$ synthetic minority observations. We compare four procedures: logistic regression fitted to the raw imbalanced data, class-weighted logistic regression with class weights $(1/2,1/2)$, naive synthetic augmentation, and synthetic augmentation with bias correction. All classifiers use linear logistic regression with an intercept and binary cross-entropy loss. Performance is evaluated using an independent balanced test sample containing 10,000 observations from each class. For each value of $\gamma$, results are averaged over 200 independent replicates.

The data-generating configuration was selected through a prespecified calibration study without using predictive performance of the bias-corrected estimator. Let $f_{\rm cal}$ denote the fixed reference logistic prediction function used in that calibration study, held fixed across values of $\gamma$, and define the pointwise calibration diagnostic
$$
D_{\rm BT}^{\rm cal}(\gamma)
=
|\Delta_{1,\gamma}(f_{\rm cal})-\Delta_0(f_{\rm cal})|.
$$
This diagnostic evaluates bias transfer at $f_{\rm cal}$ and is not an estimate of the uniform constant $\varepsilon_{\rm BT}$ in Assumption~\ref{asmp:bias_transfer}. For brevity, write $\widehat\Delta_0^{\rm cal}=\widehat\Delta_0(f_{\rm cal})$. At $\gamma=0$, the class-specific synthetic biases evaluated at $f_{\rm cal}$ were closely matched, $\Delta_0(f_{\rm cal})=0.05709$ and $\Delta_{1,0}(f_{\rm cal})=0.05577$,
giving $D_{\rm BT}^{\rm cal}(0)=0.00132$.
The majority correction at the reference function was also accurately estimated, with $E(\widehat\Delta_0^{\rm cal})=0.05606$
and $\operatorname{SD}(\widehat\Delta_0^{\rm cal})/|\Delta_0(f_{\rm cal})| = 0.192.$
As $\gamma$ increased, $\Delta_0(f_{\rm cal})$ remained at $0.05709$, whereas $\Delta_{1,\gamma}(f_{\rm cal})$ decreased to $0.01545$ at $\gamma=1$, and the corresponding pointwise discrepancy increased to $D_{\rm BT}^{\rm cal}(1)=0.04165$. The correction-noise ratio remained between approximately $0.17$ and $0.20$ across the considered values of $\gamma$.

Balanced test log loss shows the clearest effect of bias correction. At $\gamma=0$, naive synthetic augmentation yielded an average log loss of $0.42589$, whereas bias correction reduced it to $0.41858$. The paired difference between synthetic augmentation and bias correction was $0.00731$ with a 95\% confidence interval of $(0.00691,0.00771).$
As the bias-transfer discrepancy increased, the advantage of bias correction generally decreased. The paired Synthetic--BC log-loss difference declined from $0.00731$ at $\gamma=0$ to $0.00592$ at $\gamma=0.25$, $0.00519$ at $\gamma=0.5$, and $0.00408$ at $\gamma=1$. Bias correction nevertheless retained lower balanced log loss than naive synthetic augmentation over the full range considered. At $\gamma=0$, the corresponding AUCs for the raw, weighted, synthetic, and bias-corrected procedures were $0.88471$, $0.88829$, $0.88758$, and $0.88844$, respectively.

\begin{table}[t]
\centering
\setlength{\tabcolsep}{5.5pt}
\caption{Summary of the main results from the bias-transfer sensitivity simulation. The noise ratio is $\operatorname{SD}(\widehat\Delta_0^{\rm cal})/|\Delta_0(f_{\rm cal})|$. Empirical balanced-test losses for synthetic augmentation and bias correction are reported across values of $\gamma$. Positive values of ``Synthetic$-$BC'' indicate that bias correction achieves lower balanced test log loss than naive synthetic augmentation.}
\label{tab:bias_transfer_sensitivity}
\small
\begin{tabular}{cccccccc}
\toprule
$\gamma$& $D_{\rm BT}^{\rm cal}$ & $E(\widehat\Delta_0^{\rm cal})$ & Noise ratio & Synthetic & BC &  Synthetic$-$BC & 95\% CI\\
\midrule
0 & 0.00132 & 0.05606 & 0.192 & 0.42589 & 0.41858 & 0.00731 & (0.00691, 0.00771) \\
0.05& 0.00140 & 0.05581 & 0.187 & 0.42082 & 0.41392 & 0.00690 & (0.00651, 0.00729) \\
0.10& 0.00255 & 0.05612 & 0.187 & 0.41710 & 0.41015 & 0.00695 & (0.00657, 0.00733) \\
0.15 & 0.00687 & 0.05531 & 0.186 & 0.41372 & 0.40695 & 0.00677 & (0.00638, 0.00716) \\
0.20& 0.00721 & 0.05625 & 0.202 & 0.40974 & 0.40332 & 0.00641 & (0.00602, 0.00681) \\
0.25& 0.00897 & 0.05588 & 0.174 & 0.40605 & 0.40013 & 0.00592 & (0.00555, 0.00628) \\
0.50& 0.01739 & 0.05540 & 0.195 & 0.39427 & 0.38908 & 0.00519 & (0.00476, 0.00562) \\
0.75& 0.02671 & 0.05740 & 0.178 & 0.38583 & 0.38185 & 0.00398 & (0.00356, 0.00440)\\
1   & 0.04165 & 0.05578 & 0.190 & 0.38089 & 0.37680 & 0.00408 & (0.00367, 0.00449) \\
\bottomrule
\end{tabular}

\end{table}

These results illustrate the empirical role of pointwise bias transfer. When the two class-specific synthetic biases at $f_{\rm cal}$ are closely matched and the majority correction is estimated with relatively small variability, bias correction improves upon na\"{i}ve synthetic augmentation in terms of balanced test log loss. As the minority synthetic bias at this reference function moves away from the majority synthetic bias, this improvement becomes smaller, while the variability of the estimated majority correction remains relatively stable. Because $D_{\rm BT}^{\rm cal}$ is pointwise, this experiment is a sensitivity diagnostic and does not directly verify the uniform condition in Assumption~\ref{asmp:bias_transfer}.

\subsection{Average Treatment Effect Estimation}\label{suppsec_ATE_simulation}
\begin{figure}[htbp]
\centering 
\includegraphics[width=0.8\textwidth]{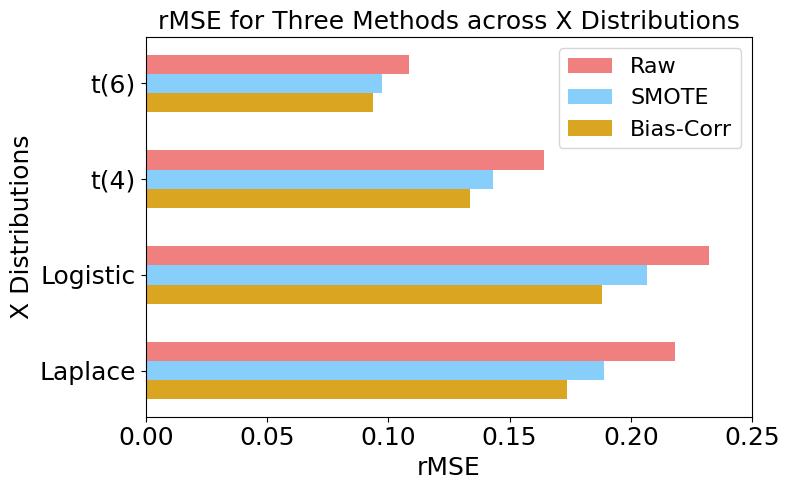}
\caption{Root mean squared error (RMSE) for ATE estimation using three methods across four covariate distributions.}
\label{simulationATE}
\end{figure}

In this section, we conduct a simulation study to evaluate the performance of three methods for estimating the average treatment effect (ATE) under covariate imbalance and distributional heterogeneity.
For each run, we generate $n=2000$ observations with $d=2$ covariates $\bm X$, and assign a binary treatment $Z$ according to a logistic propensity model
\[
\mathbb{P}(Z=1\mid \bm X)=\sigma(\beta_{z,0}+\bm \beta_z^\top \bm X),
\]
with $\beta_{z,0}=-1.6$ and $\bm \beta_z=\mathbf{1}_d$, which induces an imbalanced treated proportion.
Potential outcomes follow linear models,
\[
Y(1)=a_1+\bm X^\top \bm \mu_1+\varepsilon_1,\qquad
Y(0)=a_0+\bm X^\top \bm \mu_0+\varepsilon_0,
\]
with $a_1=a_0=1$, $\bm \mu_1=(2,1)^\top$, $\bm \mu_0=\bm \mu_1$, and independent Gaussian noise with standard deviation $\sigma_y=1$. The observed outcome is $Y=ZY(1)+(1-Z)Y(0)$, and the true ATE is $\tau=\mathbb{E}[Y(1)-Y(0)]$.
We compare (i) the standard AIPW estimator computed on the raw data, (ii) an AIPW estimator that fits the propensity model after augmenting the treated group with SMOTE-generated synthetic covariates, and (iii) the same SMOTE augmentation combined with the proposed bias-corrected propensity fitting.
The number of synthetic treated samples is chosen to target a treated share of $0.5$, and SMOTE uses $k=5$ nearest neighbors.
Covariates $X$ are generated from four distributions: $t(4)$, $t(6)$, logistic, and Laplace.
Performance is summarized by the RMSE of $\widehat\tau$ over $100$ independent runs, and the results are reported in Figure~\ref{simulationATE}.

In the four settings shown in Figure~\ref{simulationATE}, both synthetic-data procedures have lower RMSE than the raw-data implementation, and bias correction has lower RMSE than uncorrected SMOTE augmentation. These results provide a numerical illustration of how generator-induced distortion in propensity-score fitting can propagate to ATE estimation; they do not imply a uniform ordering across all data-generating mechanisms.

\subsection{MNIST Dataset}
\begin{figure}[htbp]
\centering 
\includegraphics[width=0.8\linewidth]{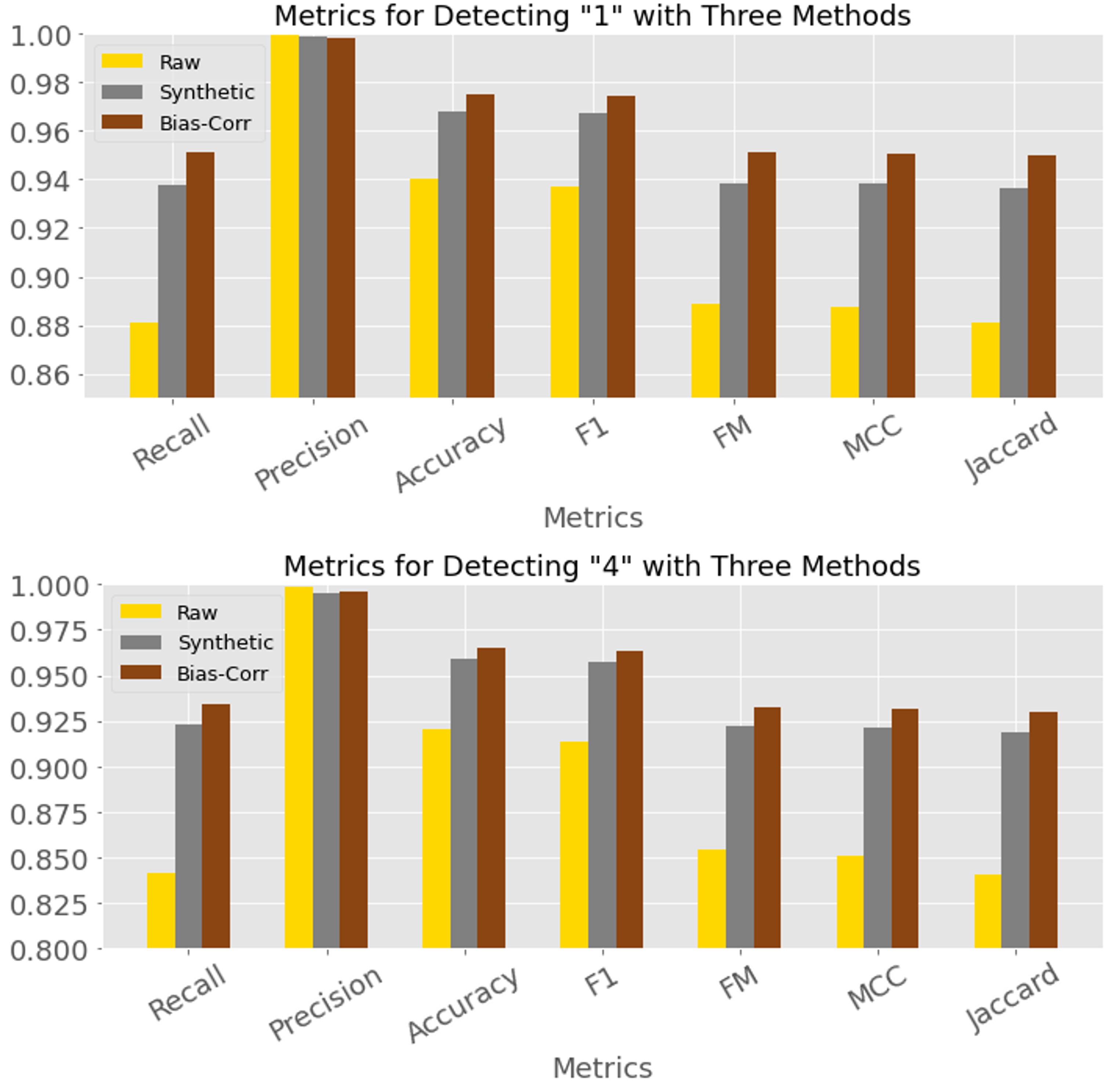}
\caption{Seven metrics (Recall, Precision, Accuracy, Fowlkes-Mallows score (FM), F1-score, Matthews Correlation Coefficient (MCC), and Jaccard index) for the three methods applied to the MNIST dataset (digits 0-4). Results are shown for binary classification of digit 1 (top panel) and digit 4 (bottom panel), treated as minority classes.}
\label{realdata2}
\end{figure}

To evaluate the practical efficacy of our proposed framework, we apply it to the MNIST dataset \citep{lecun1998mnist,mnist554dataset}. We use perturbed sampling (see Section~\ref{suppsec_syn} for its generating mechanism) to generate synthetic data for a binary classification task where digit 1 or 4 is treated as the minority class in a five-digit subset. The results are detailed in Figure~\ref{realdata2}.

In the two MNIST tasks in Figure~\ref{realdata2}, perturbed sampling improves most metrics relative to training on the raw imbalanced data while giving similar precision. Adding bias correction produces a further improvement across the reported metrics. These results are consistent with the presence of appreciable generator-induced distortion in these two tasks and illustrate how the correction can complement perturbed sampling.

\section{Synthetic Generators}\label{suppsec_syn}

We briefly review some synthetic generating methods in this section. 

\paragraph{Reweighting and Bootstrap.} Reweighting is an intuitive oversampling technique used to address imbalanced data in machine learning. This approach works by assigning a higher weight to samples from the minority class so that the training process focuses more on learning from the underrepresented group. For instance, consider a dataset with $n_1$ minority samples and $n_0$ majority samples with $n_1 \ll n_0$. A common reweighting approach assigns a weight of $w_1 = \lfloor n_0/n_1 \rfloor$ to each minority sample and a weight of $w_0 = 1$ to each majority sample \citep{breiman2017classification}. This approach is equivalent to oversampling the minority class by replicating each minority sample $\lfloor n_0/n_1\rfloor - 1$ times and training on the resulting augmented dataset with equal weight.

In contrast, bootstrap methods \citep{efron1994introduction} for imbalanced classification generate synthetic samples by randomly drawing with replacement from the minority samples. Bootstrap can be regarded as a generalization of the fixed-weight reweighting approach as it effectively assigns random weights to the minority samples in each resampling step. While both reweighting and bootstrap are intuitive and straightforward to implement, they are sensitive to outliers in the minority class. By heavily emphasizing or replicating the outliers, these approaches can potentially lead to overfitting to the noise present in the minority group.

\paragraph{Perturbed Sampling.} Perturbed sampling generates synthetic minority samples by adding small random perturbations to observed minority samples. Given minority samples $\bm X_1, \ldots, \bm X_{n_1}$, a synthetic sample is obtained as $\tilde{\bm X} = \bm X_I + \bm\epsilon$, where $I$ is drawn uniformly from $\{1, \ldots, n_1\}$ and $\bm\epsilon$ is an independent mean-zero noise vector, for example $\bm\epsilon\sim\mathcal N(\bm 0, \sigma^2\bm I_d)$ with a perturbation scale $\sigma$ chosen relative to the spread of the minority samples. The procedure can be viewed as a smoothed version of the bootstrap: instead of replicating observed samples exactly, it spreads probability mass in a neighborhood of each observed sample, which mitigates exact duplication at the cost of an additional tuning parameter. See \citet{he2009learning} for a review of perturbation-based sampling schemes for imbalanced data. This generator is used in the MNIST analysis in Section~\ref{sec_simu}.

\paragraph{Gaussian Model.} A Gaussian model is a parametric oversampling technique that approximates the minority distribution by a multivariate Gaussian distribution; it can also be viewed as the one-component special case of a Gaussian mixture model \citep{mclachlan2000finite}. Given minority samples $\bm X_1, \dots, \bm X_{n_1}$, this approach estimates the empirical mean $\widehat{\bm\mu}_1 = n_1^{-1}\sum_{i=1}^{n_1} \bm X_i$ and sample covariance matrix $\widehat{\bm\Sigma}_1 = (n_1 - 1)^{-1}\sum_{i=1}^{n_1}(\bm X_i - \widehat{\bm\mu}_1)(\bm X_i - \widehat{\bm\mu}_1)^\top$. Synthetic samples are then drawn from $\mathcal N(\widehat{\bm\mu}_1,\widehat{\bm\Sigma}_1)$. This procedure captures the first two empirical moments but can poorly represent a minority distribution that is multimodal, heavy tailed, skewed, or supported on a non-convex set.

\paragraph{Synthetic Minority Oversampling TEchnique (SMOTE).} SMOTE, introduced by \citet{chawla2002smote}, is a widely used oversampling method that generates synthetic minority samples in imbalanced datasets. SMOTE generates new synthetic samples by linearly interpolating between pairs of minority samples. It works as follows: for a randomly selected minority class sample, first find its $K$ nearest neighbors in the minority group. Then randomly select one of these $K$ nearest neighbors and create a new point along the line segment between the original point and the chosen neighbor. This procedure is repeated until the desired number of synthetic samples is reached. SMOTE requires a hyperparameter $K$, the number of nearest neighbors considered for each minority sample. Algorithm~\ref{algo_SMOTE} provides a step-wise description on how SMOTE generates $\tilde{n}_1$ synthetic samples based on input data $\bm X_1, \ldots, \bm X_{n_1}$.

\begin{algorithm}[htbp]
\caption{Synthetic Minority Oversampling TEchnique (SMOTE)}
\label{algo_SMOTE}
\begin{algorithmic}[1]
    \INPUT Samples $(\bm X_i)_{i=1}^{n_1}$, the number of nearest neighbors $K$, synthetic sample size $\tilde{n}_1$.    
    \FOR{each $i$ in $1:n_1$}
        \STATE Find the $K$ nearest neighbors of $\bm X_i$, denoted as $\bm X_{i(1)}, \ldots, \bm X_{i(K)}$.
    \ENDFOR
    \FOR{each $i$ in $1:\tilde{n}_1$}
        \STATE Sample index $t$ uniformly from $\{1, 2, \dots, n_1\}$.\label{algo_sample_t}
        \STATE Sample $U_i$ from $\mathcal U(0, 1)$, i.e., from the uniform distribution on the interval [0, 1]. 
        \STATE Sample $k$ uniformly from $\{1, \ldots, K\}$.
        \STATE Generate the SMOTE sample $\tilde{\bm X}_i^{(1)} \leftarrow \bm X_t + U_i(\bm X_{t(k)} - \bm X_t)$.
    \ENDFOR
    \OUTPUT Synthetic samples $(\tilde{\bm X}_i^{(1)})_{i=1}^{\tilde{n}_1}$.
\end{algorithmic}
\end{algorithm}

\paragraph{Diffusion Model.} Diffusion models \citep{ho2020denoising, song2020score} form one of the most popular classes of generative models for data synthesis. A diffusion model learns the distribution of observed samples by simulating and statistically revising a Markovian diffusion process that maps the data to standard Gaussian noise and then reconstructs data from noise. The framework consists of two phases: a fixed forward process, which maps a data example to Gaussian noise, and a learned backward process, which iteratively maps random noise back to a realistic data sample. 

The forward process is a fixed Markov chain that progressively corrupts a sample with Gaussian noise over $T$ time steps, parameterized by a schedule of variance terms $\beta_t \in (0, 1)$ for $t = 1, \dots, T$. Starting with an original data sample $\bm x\in\mathbb R^d$ and letting $\bm z_0 = \bm x$, a series of intermediate latent variables $\bm z_1, \dots, \bm z_T\in\mathbb R^d$ are generated according to the following iterative equation,
\begin{align*}
    \bm z_t =&\: \sqrt{1-\beta_t}\cdot \bm z_{t-1} + \sqrt{\beta_t}\cdot\bm\epsilon_t, \quad t = 1,\dots, T,
\end{align*}
where $\bm\epsilon_t \sim \mathcal N(\bm 0, \bm I_d)$ is noise added at time $t$. Denoting $\alpha_t = \prod_{s=1}^t(1- \beta_s)$ for $t = 1, \dots, T$, this process allows for a direct-sampling property, which makes it possible to obtain $\bm z_t$ from $\bm x$ in one step:
\begin{align*}
    \bm z_t = \sqrt{\alpha_t}\cdot \bm x + \sqrt{1-\alpha_t}\cdot \bm\epsilon,
\end{align*}
where $\bm\epsilon \sim \mathcal N(\bm 0, \bm I_d)$. Since $\beta_t < 1$ is chosen such that $\alpha_T \approx 0$ for large $T$, the final latent variable $\bm z_T$ is guaranteed to be close to the standard Gaussian distribution $\mathcal N(\bm 0, \bm I_d)$. 

The backward process defines the generative model. It defines a learned Markov chain that attempts to reverse the diffusion process, starting from pure noise $\bm z_T \sim \mathcal N(\bm 0, \bm I_d)$ and iteratively denoising it back to a data sample $\bm z_0$:
\begin{align*}
    \bm z_{t-1}\mid (\bm z_t, \phi_t) \sim \mathcal N(\bm f_t(\bm z_t, \phi_t), \sigma_t^2\bm I), \quad t = T, T-1, \dots, 1.
\end{align*}
The function $\bm f_t(\bm z_t, \phi_t)$ is a neural network that is trained to estimate the mean of the approximate Gaussian distribution for the mapping from $\bm z_t$ to $\bm z_{t-1}$, and $\sigma_t$ is predetermined by the variance parameter $\beta_t$. By chaining these steps, diffusion models can synthesize high-fidelity data by gradually transforming Gaussian noise to structured samples.

\paragraph{Flow matching.} Flow matching \citep{lipman2022flow} aims to learn a smooth and invertible map from a simple base distribution, say, the standard Gaussian distribution, to the target data distribution. For an observation $\bm x\in\mathbb R^d$, consider the probability density path $p: [0, 1] \times \mathbb R^d \to \mathbb R_+$ such that $\int p_t(\bm x){\rm d}\bm x = 1$ for any $t\in[0, 1]$. Let $p_0$ be the simple base distribution and $p_1$ the target data distribution. Define the flow $\bm\phi: [0, 1] \times \mathbb R^d \to \mathbb R^d$ as a time-dependent diffeomorphic map such that if $\bm x \sim p_0$, then $\bm\phi_t(\bm x) \sim p_t$. Without loss of generality, let $\bm\phi_0(\bm x) = \bm x$. The flow can be generated by a continuous normalizing-flow vector field $\bm v: [0, 1] \times \mathbb R^d \to \mathbb R^d$ satisfying $\frac{\rm d}{{\rm d}t}\bm\phi_t(\bm x) =\bm v_t(\bm\phi_t(\bm x))$ \citep{chen2018neural}. Flow matching simplifies the learning problem by using conditional flows that define a straight path between noise and a data point. For a given data point $\bm x_1 \sim p_1$ and noise sample $\bm x_0 \sim p_0$, the optimal conditional vector field is the straight-line path $\bm u_t(\bm x) = \bm x_1 - \bm x_0$.

The goal of flow matching is to train a neural network field $\bm v_t(\bm x; \theta)$ parameterized by $\theta$ to match the ideal conditional vector field $\bm u_t$ in expectation. Suppose the target probability density path $p_t$ is generated by the vector field $\bm u_t$, flow matching aims to minimize the objective function
\begin{align*}
    L_{\rm FM}(\theta) = \mathbb E_{t, p_0, p_1}\left[\|v_t(\bm x_0 + t(\bm x_1 - \bm x_0); \theta) - (\bm x_1 - \bm x_0)\|^2\right],
\end{align*}
where $t\sim U(0, 1)$, $\bm x_0 \sim p_0$ and $\bm x_1 \sim p_1$.
With the learned vector field $\bm v_t(\bm x; \theta)$ and a random noise sample $\bm z\sim p_0$, synthetic samples are generated by
\begin{align*}
    \tilde{\bm x} = \bm\phi_1(\bm z), \quad \text{where } \frac{\rm d}{{\rm d}t}\bm\phi_t(\bm z) = \bm v_t(\bm\phi_t(\bm z); \theta).
\end{align*}

There are many other synthetic generators, such as generative adversarial networks (GANs) \citep{goodfellow2014generative, goodfellow2020generative}, normalizing flows \citep{rezende2015variational}, and variational autoencoders (VAE) \citep{kingma2013auto}. Please see \citet{figueira2022survey, lu2023machine} for a comprehensive survey.
\end{sloppypar}

\end{document}